\documentclass[default,oneside]{sn-jnl}

\theoremstyle{thmstyleone}%
\usepackage[numbers,sort&compress]{natbib}
\setcitestyle{open={[},close={]}}

\theoremstyle{thmstyletwo}%
\theoremstyle{thmstylethree}%

\usepackage{bm}
\usepackage{pifont}
\usepackage{subcaption}
\usepackage{graphicx}%
\usepackage{geometry}
\usepackage{multirow}%
\usepackage{amsmath,amssymb,amsfonts}%
\usepackage{amsthm}%
\usepackage{mathrsfs}%
\usepackage[title]{appendix}%
\usepackage{textcomp}%
\usepackage{manyfoot}%
\usepackage{booktabs}%
\usepackage{algorithm}%
\usepackage{algorithmicx}%
\usepackage{algpseudocode}%
\usepackage{listings}%
\usepackage{soul}
\usepackage{multirow}
\usepackage{graphicx}
\usepackage{float}
\usepackage{makecell} 
\usepackage{indentfirst}
\usepackage{hyperref}
\usepackage{setspace}
\usepackage{tabularx} 
\usepackage{ragged2e} 
\usepackage{arydshln}
\newcolumntype{L}{>{\RaggedRight\hangafter=1\hangindent=0em}X}

\jyear{2023}%

\theoremstyle{thmstyleone}%
\theoremstyle{thmstyletwo}%

\theoremstyle{thmstylethree}%

\makeatletter

\newcommand{\Rmnum}[1]{\expandafter\@slowromancap\romannumeral #1@}
\makeatother

\makeatletter
	\def\@cite#1#2{\textsuperscript{{#1\if@tempswa , #2\fi}}}
\makeatother

\usepackage{etoolbox}
\makeatletter
\patchcmd{\ps@headings}
{\hbox to \hsize{\hfill Springer Nature 2021 \LaTeX\ template\hfill}}
{\hbox to \hsize{}}
{}
{}
\patchcmd{\ps@titlepage}
{\hbox to \hsize{\hfill Springer Nature 2021 \LaTeX\ template\hfill}}
{\hbox to \hsize{}}
{}
{}
\makeatother
\makeatletter
\patchcmd{\ps@headings}
{\hbox to \hsize{\hfill Springer Nature 2021 \LaTeX\ template\hfill}}
{\hbox to \hsize{}}
{}
{}

\makeatother

\begin{document}


\title[AIMatDesign]{\textsc{AIMatDesign}: Knowledge-Augmented Reinforcement Learning for Inverse Materials Design under Data Scarcity}


\author[1]{\fnm{Yeyong} \sur{Yu}}\email{yuyeyong@shu.edu.cn}

\author[2]{\fnm{Xilei} \sur{Bian}}\email{bianxilei@shu.edu.cn}

\author[2]{\fnm{Jie} \sur{Xiong}}\email{xiongjie@shu.edu.cn}

\author[1,3,4]{\fnm{Xing} \sur{Wu}}\email{xingwu@shu.edu.cn}

\author*[1,2,3,4]{\fnm{Quan} \sur{Qian}}\email{qqian@shu.edu.cn}

\affil[1]{\orgdiv{School of Computer Engineering \& Science}, \orgname{Shanghai University}, \orgaddress{ \city{Shanghai}, \postcode{200444}, \country{China}}}

\affil[2]{\orgdiv{Center of Materials Informatics and Data Science, Materials Genome Institute},
	\orgname{Shanghai University}, \orgaddress{ \city{Shanghai}, \postcode{200444}, \country{China}}}

\affil[3]{\orgdiv{Key Laboratory of Silicate Cultural Relics Conservation (Shanghai University)}, \orgname{Ministry of Education}, \country{China}}

\affil[4]{\orgdiv{Shanghai Institute for Advanced Communication and Data Science},
	\orgname{Shanghai University}, \orgaddress{ \city{Shanghai}, \postcode{200444}, \country{China}}}
            


\abstract{With the growing demand for novel materials, machine learning-driven inverse design methods face significant challenges in reconciling the high-dimensional materials composition space with limited experimental data. 
Existing approaches suffer from two major limitations:
(\uppercase\expandafter{\romannumeral1}) machine learning models often lack reliability in high-dimensional spaces, leading to prediction biases during the design process;
(\uppercase\expandafter{\romannumeral2}) these models fail to effectively incorporate domain expert knowledge, limiting their capacity to support knowledge-guided inverse design. 
To address these challenges, we introduce \textsc{AIMatDesign}, a reinforcement learning framework that addresses these limitations by augmenting experimental data using difference-based algorithms to build a trusted experience pool, accelerating model convergence. 
To enhance model reliability, an automated refinement strategy guided by large language models (LLMs) dynamically corrects prediction inconsistencies, reinforcing alignment between reward signals and state value functions. 
Additionally, a knowledge-based reward function leverages expert domain rules to improve stability and efficiency during training. 
Our experiments demonstrate that \textsc{AIMatDesign} significantly surpasses traditional machine learning and reinforcement learning methods in discovery efficiency, convergence speed, and success rates.
Among the numerous candidates proposed by \textsc{AIMatDesign}, experimental synthesis of representative Zr-based alloys yielded a top-performing BMG with 1.7GPa yield strength and 10.2\% elongation, closely matching predictions.
Moreover, the framework accurately captured the trend of yield strength variation with composition, demonstrating its reliability and potential for closed-loop materials discovery.
This approach provides an innovative solution for efficient inverse materials design, opening promising avenues for intelligent materials development under data-limited conditions.}



\maketitle

\section{Introduction}
\label{Introduction}

The accelerating demand for rapid design and discovery of novel materials is propelling computationally-driven materials research into new frontiers. 
Traditional experimental approaches relying on iterative trial-and-error are time-consuming, labor-intensive, and cost-prohibitive, limiting their ability to meet the requirements of fast-paced materials design and iterative optimization. 
Given the high-dimensional complexity of material composition spaces and associated performance characteristics, inverse design methodologies are increasingly adopting intelligent exploration approaches based on artificial intelligence (AI). 

\begin{figure}[h]
    \centering
    \includegraphics[width=\textwidth]{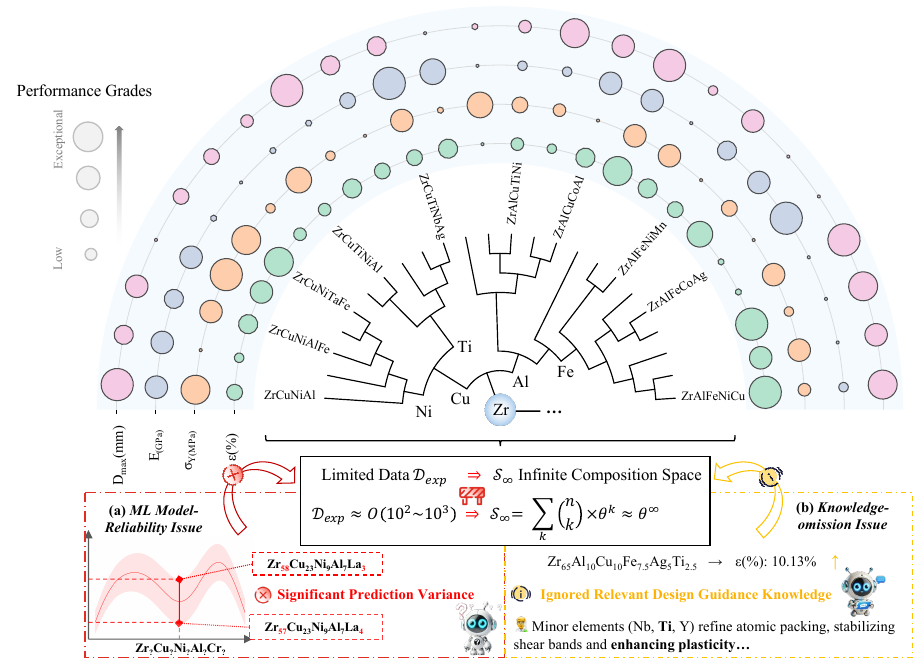}
    \caption{Sparse sampling of the vast materials composition space reveals two recurrent issues in current machine learning–guided inverse materials design. 
    (a) \textit{Model‑reliability Issue}: compositions clustered in the red box exhibit large prediction variance despite near‑identical chemistries, highlighting the brittleness of static surrogate models. 
    (b) \textit{Knowledge‑omission Issue}: the elongation $\varepsilon$ (\%) of Ti‑containing Zr$_{65}$Al$_{10}$Cu$_{10}$Fe$_{7.5}$Ag$_{5}$Ti$_{2.5}$ is systematically under‑predicted (yellow box) because Ti‑induced plasticity is not encoded in the training data.}
    \label{fig:intro}
\end{figure}

However, there remains a significant gap between limited experimental data $\mathcal{D}_{\text{exp}} \approx O(10^2 \sim 10^3)$ and the practically infinite composition space $\mathcal{S}_{\infty} \approx \theta^{\infty}$\footnote{The full expression of the composition space is $\mathcal{S}_{\infty} = \sum_k \binom{n}{k} \times \theta^k$, where $n$ is the number of possible constituent elements, $k$ is the number of components in a composition, and $\theta$ denotes the exploration range for each component.}.
As illustrated in Fig.~\ref{fig:intro}, the existing dataset of \textit{Zr-Based Bulk Metallic Glasses (BMGs)} occupies an extremely limited fraction of the vast possible composition space. 
Further analysis reveals that current machine learning (ML)-guided inverse design methods primarily face two critical challenges:
\begin{itemize}
    \item \textbf{Model-reliability Issue}.  
          Static surrogate models trained on finite datasets cannot adaptively correct bias or noise during optimisation.  
          This limitation manifests as the high‑variance cluster marked by the \textcolor{red}{\textbf{red box}} in Fig.~\ref{fig:intro}\,(a).
    \item \textbf{Knowledge‑omission Issue}.  
          Purely data‑driven pipelines overlook mechanistic insights that domain experts routinely exploit—e.g.\ Ti additions are known to enhance plasticity in Zr‑based BMGs.  
          The \textcolor{orange}{\textbf{yellow box}} in Fig.~\ref{fig:intro}\,(b) illustrates how the absence of such priors leads to systematic under‑prediction of elongation.
\end{itemize}

Due to $\mathcal{D}_{\text{exp}} \ll \mathcal{S}_{\infty}$, existing ML-driven approaches struggle to achieve stable generalization under sparse data conditions. 
Furthermore, the absence of expert domain knowledge severely constrains the model's exploration efficiency, often leading to suboptimal outcomes. 
These two issues further highlight the inherent difficulty of $\mathcal{D}_{\text{exp}} \Rightarrow \mathcal{S}_{\infty}$, revealing the limitations of current ML-driven inverse design methods in efficiently exploring novel materials within data-scarce conditions.

To address these challenges and enhance AI-driven inverse materials design, we propose a innovative reinforcement learning (RL)-based framework named-\textsc{AIMatDesign}. 
Compared with conventional methods, RL offers strong adaptability and dynamic decision-making, enabling step-by-step exploration of optimal solutions in high-dimensional, complex design spaces through iterative interaction with the environment. 
Specifically, we employ a difference-based strategy to augment limited experimental data into a large, Trustworthy Experience Pool (TEP) for RL training, effectively addressing the data scarcity issue.
We further introduce a Knowledge-Based Reward (KBR) system and an Automatic Model Refinement (AMR) strategy to improve the model’s decision-making capabilities, ensuring efficient and accurate exploration within the extensive composition space ($\mathcal{S}_{\infty}$).

We applied \textsc{AIMatDesign} to the BMGs design task to evaluate its experimental performance. 
As demonstrated in \textsection~\ref{rl_results}, \textsc{AIMatDesign} achieves notable improvements in both convergence speed and success rate for inverse materials design compared to traditional optimization methods (e.g., grid search and NSGA-II) as well as other mainstream RL baselines. 
These results confirm the feasibility and advantages of \textsc{AIMatDesign} in complex materials design tasks, providing an innovative and efficient pathway for AI-driven inverse design.
Our main contributions are summarized as follows:
\begin{itemize}
	\item We developed a RL-based framework for efficiently exploring high-dimensional materials composition spaces ($\mathcal{S}_{\infty}$), integrating an adaptive reward mechanism to effectively guide inverse design. To overcome the limitations of scarce experimental data ($\mathcal{D}_{\text{exp}}$), we employed a difference-based strategy to expand the limited $\mathcal{D}_{\text{exp}}$ into a \textbf{Trustworthy Experience Pool (TEP)}, facilitating rapid RL model convergence within the extensive space ($\mathcal{S}_{\infty}$) through a progressive guidance strategy.
	\item To address reliability issues commonly faced by ML models in inverse design, we proposed two \textbf{Automatic Model Refinement (AMR)} strategies: \textit{Variance-Based Refinement} and \textit{Correlation-Based Refinement}. When reliability deviations are detected, LLMs are employed to automatically refine ML predictions, enhancing consistency between reward signals and state value functions, significantly improving the stability and convergence efficiency of RL.
	\item To bridge the gap created by the absence of expert knowledge in purely data-driven approaches, we innovatively integrated domain-specific materials knowledge into the inverse design process via LLMs. By leveraging a \textbf{Knowledge-Based Reward (KBR)} strategy at critical stages, we effectively combined data-driven predictions with expert insights, substantially enhancing the overall accuracy and efficiency of materials inverse design.
	\item Guided by our framework, we successfully discovered novel Zr-Based BMGs, with experimental validation confirming a yield strength of up to 1.7GPa and 10.2\% elongation—closely aligned with predictions—highlighting the framework’s practical effectiveness in closed-loop materials discovery.
\end{itemize}

\section{Related Work}
\label{Related Work}

\noindent \textbf{Conventional Paradigms in Inverse Materials Design} The paradigm of inverse materials design has evolved from experimental-driven to theory-driven and, more recently, to computation-driven approaches. Each paradigm has made unique contributions under different research contexts, while also exhibiting inherent limitations.
\begin{enumerate}
	\item[(1)] The \textbf{experimental-driven approach} primarily relies on trial-and-error strategies, identifying promising materials through accumulated empirical data. While this method allows direct verification of material properties, its high cost, long development cycles, and limited exploration scope severely constrain its broader application~\citep{field2001market, capjon2004trial}.
	\item[(2)] \textbf{Theory-driven methods}, such as density functional theory (DFT)~\citep{kang2022advances, saal2013materials} and high-throughput simulations~\citep{maier2007combinatorial}, provide atomic-scale property predictions and reduce experimental demands. However, their high computational cost and limited scalability constrain their use in complex systems.
	\item[(3)] \textbf{Computation-driven strategies}, including Monte Carlo Tree Search (MCTS)~\citep{m2017mdts} and genetic algorithms~\citep{frazier2016bayesian, chakraborti2004genetic}, improve search efficiency by simulating and optimizing the design process. Still, they struggle with high-dimensional and uncertain design spaces, limiting their effectiveness in complex materials discovery.
\end{enumerate}

\noindent \textbf{Machine Learning-Driven Inverse Materials Design} 
With the growing demand for more efficient and intelligent approaches in inverse materials design, \textbf{ML-driven methods} have emerged as a promising tool for materials discovery and optimization~\citep{liu2017materials}. 
Compared to conventional paradigms, ML offers higher efficiency in data mining, allowing for rapid exploration of vast materials spaces~\citep{ramprasad2017machine}. 
Generative models such as generative adversarial networks (GANs)~\citep{menon2022generative, dan2020generative} and variational autoencoders (VAEs)~\citep{lew2021encoding, el2023vae} offer new pathways for designing novel materials by learning complex structure–property relationships through generation–discrimination or encoding–decoding schemes.
Graph neural networks (GNNs) have also shown significant advantages in representing crystal structures flexibly, enabling improved lattice analysis and property prediction~\citep{guo2020semi, wang2021inverse}. 
Moreover, to facilitate multi-objective design and performance optimization, multi-objective algorithms such as NSGA-II and Bayesian optimization have been employed to rapidly approach optimal solutions while balancing performance, cost, and manufacturability~\citep{yu2024small, zhang2021multi, ma2023comprehensive}. 
However, the scarcity and imbalance of materials data, combined with the limited integration of expert knowledge, continue to pose challenges for model generalization, robustness, and interpretability.

\noindent \textbf{Reinforcement Learning-Driven Inverse Materials Design} 
As ML-driven inverse materials design matures, RL has attracted increasing attention as an intelligent decision-making method capable of adaptive exploration and optimization in complex strategy spaces~\citep{li2017deep}. 
Compared with approaches based on generative models or multi-objective optimization algorithms, RL enables efficient searches over discrete materials spaces through a dynamic ``trial-error–feedback–update" process, continuously refining its policy during the learning phase~\citep{sui2021deep, brown2022deep, shah2021reinforcement}. 
To address the common issue of physical constraints in materials design, researchers have integrated chemical and materials priors into the reward function, ensuring effective exploration and adherence to constraints in both discrete and continuous action spaces~\citep{karpovich2024deep}. 
However, in the absence of sufficient high-quality real-world experience to guide the process, RL models may be prone to overfitting and demonstrate limited generalization capability.

\noindent \textbf{Integration of Domain Knowledge}
With the increasing adoption of machine learning and reinforcement learning in inverse materials design, the need to integrate traditional materials science knowledge with modern data-driven approaches has become increasingly evident~\citep{jia2024llmatdesign}. 
In the inverse design process, LLMs can assist not only in generating or interpreting textual representations of material structures~\citep{liu2024beyond}, but also in candidate screening and performance evaluation, offering cross-validation that helps reduce uncertainties introduced by large-scale searches~\citep{yu2024small}. 
These efforts highlight the potential of incorporating domain knowledge into RL workflows and inspire our approach to explicitly integrate expert knowledge into the reinforcement learning loop, reinforcing its value in inverse materials design.

\section{Methods}
\label{Methods}

\begin{figure}[ht]
	\centering
	\includegraphics[width=\textwidth]{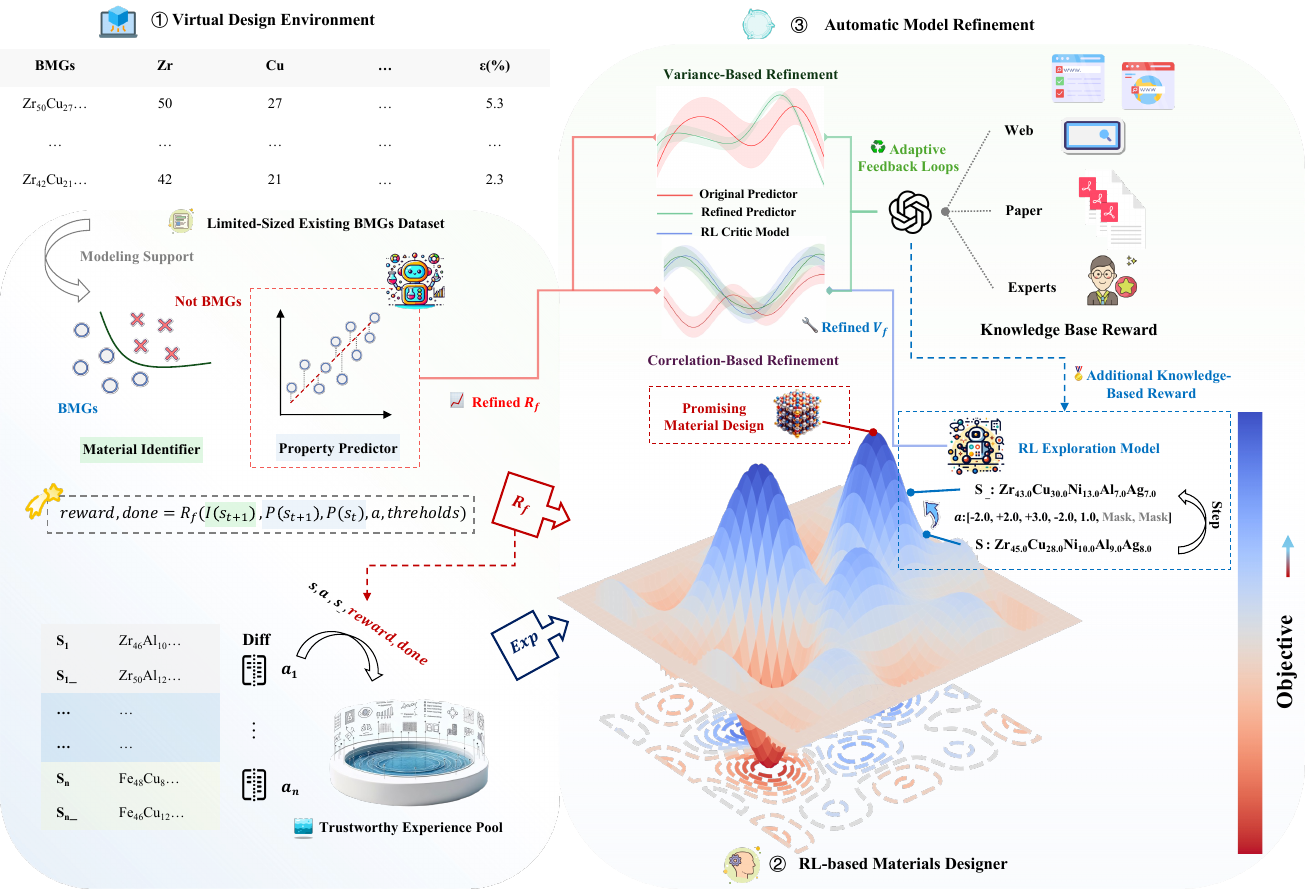}
	\caption{Schematic overview of \textsc{AIMatDesign}, with numbered components: 
	\textbf{\ding{172} Virtual Design Environment}, where ML classification and regression models plus a difference‑based experience pool define property‑centric reward functions; 
	\textbf{\ding{173} RL-based Materials Designer}, in which the agent performs additive/subtractive element manipulations across the infinite composition space under combined rewards from the simulator and LLM evaluations; 
	\textbf{\ding{174} Automatic Model Refinement}, where LLM‑derived expert knowledge iteratively corrects ML model guidance and steers RL exploration to ensure robustness.}
	\label{fig:graph_abstract}
\end{figure}

As illustrated in Fig.~\ref{fig:graph_abstract}, our proposed framework \textsc{AIMatDesign} consists of three key components:

\begin{itemize}
	\item[\ding{172}] \textbf{Virtual Design Environment.}
	We build a virtual design environment using limited-sized BMGs datasets $\mathcal{D}_{exp}$, where machine learning models for classification and regression serve as predictive guides. Reward functions are defined based on performance thresholds relevant to target properties. To improve data efficiency, a difference-based strategy is employed to extract a large set of trustworthy experience samples from the original data, forming the foundation for RL training.
	\item[\ding{173}] \textbf{RL-based Materials Designer.}
	In this environment, the RL agent explores the material composition space ($\mathcal{S}_{\infty}$) by performing additive or subtractive operations on material elements. The agent is iteratively trained using reward signals provided by both the virtual environment and LLMs.
	\item[\ding{174}] \textbf{Automatic Model Refinement.}
	To address potential reliability issues of the ML model or inconsistencies between the ML model’s guidance and the RL agent’s actions during training, LLMs—leveraging expert knowledge drawn from literature, online sources, or domain expertise—are employed to dynamically refine the ML model and correct the RL agent’s exploration path. This ensures the robustness and credibility of the design process.
\end{itemize}

\noindent Details of each component's implementation will be elaborated in the following sections. The complete training procedure is summarized in Algorithm~\ref{alg:AIMatDesign}.

\subsection{RL-Based Material Design}
\label{rl-based_material_design}

In the field of materials design, traditional optimization methods—such as Grid Search~\citep{liashchynskyi2019grid}, Bayesian optimization~\citep{shahriari2015taking}, and NSGA-II~\citep{deb2002fast}—can offer reasonable performance in low-dimensional spaces. 
However, their efficiency drops significantly when applied to high-dimensional, complex design spaces, and they often struggle to adapt to the diverse characteristics of different material systems~\citep{tian2024high, gao2019topology, liao2024topological}.
Moreover, these approaches lack intelligent self-correction mechanisms; they cannot automatically revise erroneous guidance or adjust strategies during the optimization process, which may result in convergence to suboptimal local solutions and limited exploration of the broader material space~\citep{muc2021introduction}.

To overcome these limitations, we propose a RL-based framework \textbf{\textsc{AIMatDesign}} for intelligent materials design. 
By leveraging RL’s strong exploratory capabilities in high-dimensional spaces, the framework learns to search for optimal solutions through continuous interaction and feedback. 

Unlike traditional methods, RL can progressively identify promising directions within vast design spaces and dynamically adjust decision-making strategies through environment interaction, thus significantly improving search efficiency~\citep{karpovich2024deep}.

\subsubsection{Virtual Design Environment for RL-based Materials Designer}
\label{virtual_design_environment}

In this study, we constructed \textit{Classification and Regression models} based on existing material data to provide an accurate and reliable virtual environment for \textit{RL-based Materials Designer}.

These precise machine learning models are essential for effective RL exploration, as they offer crucial guidance by accurately predicting the categories and properties of material compositions. 
Specifically, the classification model helps identify the likelihood of target materials (e.g., BMGs), thereby clarifying the optimization direction, while the regression model predicts material properties, providing a quantitative basis for the reward mechanism. 

With these high-precision predictions, reinforcement learning can receive reliable feedback in the complex materials design space, ensuring that the agent makes correct decisions during the optimization process.

\begin{figure}[ht]
	\centering
	\includegraphics[width=\textwidth]{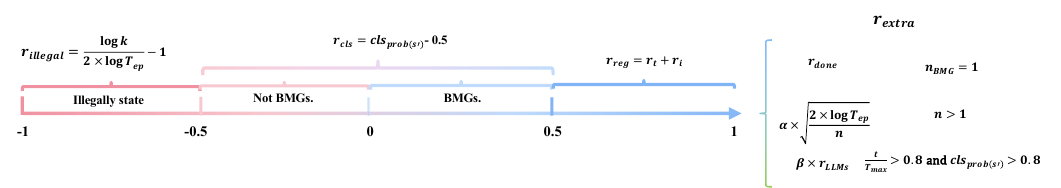}
	\caption{Hierarchical Reward Design Combining Classification, Performance Prediction, and Expert Knowledge for for Material Exploration.}
	\label{fig:reward_func}
\end{figure}

The reward function requires inputs of the original material composition $s$, the action $a$, and the resulting next state $s'$. The classification and regression models predict the categories and properties of $s$ and $s'$, which are then input into Fig.~\ref{fig:reward_func} to calculate the quantized reward $R_f$ for the pair $(s, a)$:
\begin{itemize}
	\item \textbf{Invalid State Reward}: If the action $a$ exceeds a specified threshold or leads to an invalid material composition state $s'$ (e.g., a component $< 0$ or $> 100$), the lowest reward is assigned. Specifically, $r_{\text{illegal}}$ is calculated as:
	\begin{equation}
		r_{\text{illegal}} = \frac{\log(k)}{2 \cdot \log(T_{\text{ep}})} - 1
	\end{equation}
\noindent This reward is related to the current step $k$ and the maximum number of steps $T_{\text{ep}}$ in the current round. The larger the step, the longer the agent persists in exploration, and the reward approaches -0.5. Conversely, smaller steps lead to rewards closer to -1.
	\item \textbf{Material Classification Reward}: If both $s'$ and $a$ are valid, the classification model is used to predict the probability $cls_{\text{prob}}(s')$ that $s'$ belongs to the target material type (e.g., BMG). Subtracting 0.5 gives the classification reward. The higher the classification probability for $s'$, the closer the reward is to 0.5; otherwise, it approaches -0.5.
	\item \textbf{Performance Prediction Reward}: If the probability of $s'$ belonging to the target material class is $> 0.5$, the regression model is used to predict the properties of $s'$ and calculate the performance improvement reward $r_i$ for $s'$. Additionally, the performance of $s'$ is compared to set thresholds, and the number of threshold-exceeding performances is used to assign $r_t$:
	\begin{equation}
		r_{\text{reg}}(s, s') = \underbrace{\sum_{t \in \mathcal{T}} w_t \cdot \tanh\left( 
		\frac{\hat{y}_t(s') - \hat{y}_t(s)}{\max\left\{\tau_t, \hat{y}_t(s)\right\}} 
		\right)}_{r_i}
		+ \underbrace{\sum_{t \in \mathcal{T}} w_t \cdot \mathbb{I}\left[ 
		\hat{y}_t(s') \ge \tau_t 
		\right]}_{r_t}
		\end{equation}
	\noindent where $\mathcal{T}$ is the set of target properties, $\hat{y}$ represents the predicted property values, $\tau$ is the performance threshold, and $w$ is the reward weight of each property.
	\item Beyond these three reward functions, additional rewards $\bm{r_{extra}}$ may be given if $s'$ meets specific conditions:
	\begin{itemize}
		\renewcommand{\labelitemii}{$\circ$}
		\item \textbf{New Material Reward}: If all performance thresholds are met and the material does not exist in the existing materials database, the RL model is considered to have discovered a new material, completing the current design task. In this case, the reward for that step $r_{\text{done}}$ is set to 1.
		\item \textbf{Existing Material Reward}: If $s'$ meets all performance thresholds but already exists in the materials database, $r_{\text{done}}$ is adjusted by Upper Confidence Bound 1 (UCB1)~\cite{drugan2013designing}:
		\begin{equation}
			r_{\text{done}} = \alpha \times \sqrt{\frac{2 \cdot \log(T_{\text{ep}})}{n}}
		\end{equation}
		\noindent where $n$ represents the number of times the material composition has been explored. The more often the material is explored, the more the reward decays.
		\item \textbf{Knowledge-Based Reward}: Furthermore, once 80\% of the training steps are completed and $cls_{\text{prob}}(s') > 0.8$, the LLM is used to evaluate the material composition $s'$ based on an expert knowledge base. A confidence score ranging from -1 to 1 is provided, and the RL model is rewarded accordingly with $r_{\text{LLMs}}$. \footnote{To maintain the original scale of the reward, a weight configuration $\beta$ is introduced when using the LLM reward, ensuring no unnecessary changes to the overall reward scale.}
	\end{itemize}
\end{itemize}

\subsubsection{Trustworthy Experience Pool}
\label{trustworthy_experience_pool}

To address the issue of scarce material data, we propose an innovative method for constructing a Trustworthy Experience Pool (TEP). 
This method generates a rich and reliable experience pool by computing the \textbf{differences between existing material data}. 
Specifically, we perform a differential operation on each pair of material data $s_1$ and $s_2$ in the database (i.e., $a = s_1 - s_2$), generating the corresponding action $a$ and calculating its associated reward $R_f(s_1, s_2, a)$. 
These data are then stored in the experience pool. 
Assuming there are $n$ data points, the differential operation results in $n \times (n-1)$ new experience data points, denoted as $Exp(s_1, s_2, a, r)$. 
Since these experience data directly originate from real material samples, their trustworthiness is significantly higher than data generated through classification and regression models, providing a more robust training foundation for reinforcement learning.

This differential method for constructing the experience pool not only extracts a large number of high-quality training samples from limited material data but also offers diverse exploration paths for the RL model. 
It significantly alleviates the limitations imposed by data scarcity during RL training.

To further enhance training efficiency, we introduce an experience sampling mechanism based on the mean reward of the current round. The specific strategy is as follows:

\begin{itemize}
	\item When the current round’s reward is below the TEP average, a portion of the training batch is replaced with higher-reward experiences from the TEP (exceeding the current average by more than 0.2) to strengthen learning.
	\item When the current round’s reward exceeds the TEP average, the model reduces the proportion of TEP-based replacements, relying more on real-time exploration.
\end{itemize}

This reward-based sampling strategy is essentially a progressive guiding process. 
By continuously providing high-quality training samples, the model gradually improves its exploration capabilities, avoiding premature convergence to suboptimal strategies, and achieving faster and more stable convergence.

Overall, the proposed method for constructing the TEP maximizes the potential of limited data, transforming scarce data into efficient training resources. 
This provides a strong foundation for applying reinforcement learning to materials design, helping to overcome data bottlenecks in high-dimensional, complex design spaces and significantly improving design efficiency.

\subsection{Automatic Model Refinement via Adaptive Feedback Loops}
\label{automatic_model_refinement}

Traditional materials inverse design methods, such as Bayesian optimization, primarily rely on experimental data to continuously update and improve models~\citep{shahriari2015taking}.
However, the limitation of these approaches lies in their heavy dependence on actual experimental feedback, and in cases of scarce data, the optimization efficiency is often unsatisfactory. 

In current data-driven materials inverse design workflows, machine learning models typically serve as guiding tools~\citep{liu2017materials, yu2024small, zhang2021multi}, but once the model is trained, it lacks the ability to dynamically update and self-correct, limiting its adaptability in complex environments.

Furthermore, existing inverse design methods lack effective mechanisms to timely assessing model reliability, leading to potential biases during the design process that may affect the reliability and accuracy of the final design outcomes.

\begin{figure}[ht]
	\centering
	\includegraphics[width=\textwidth]{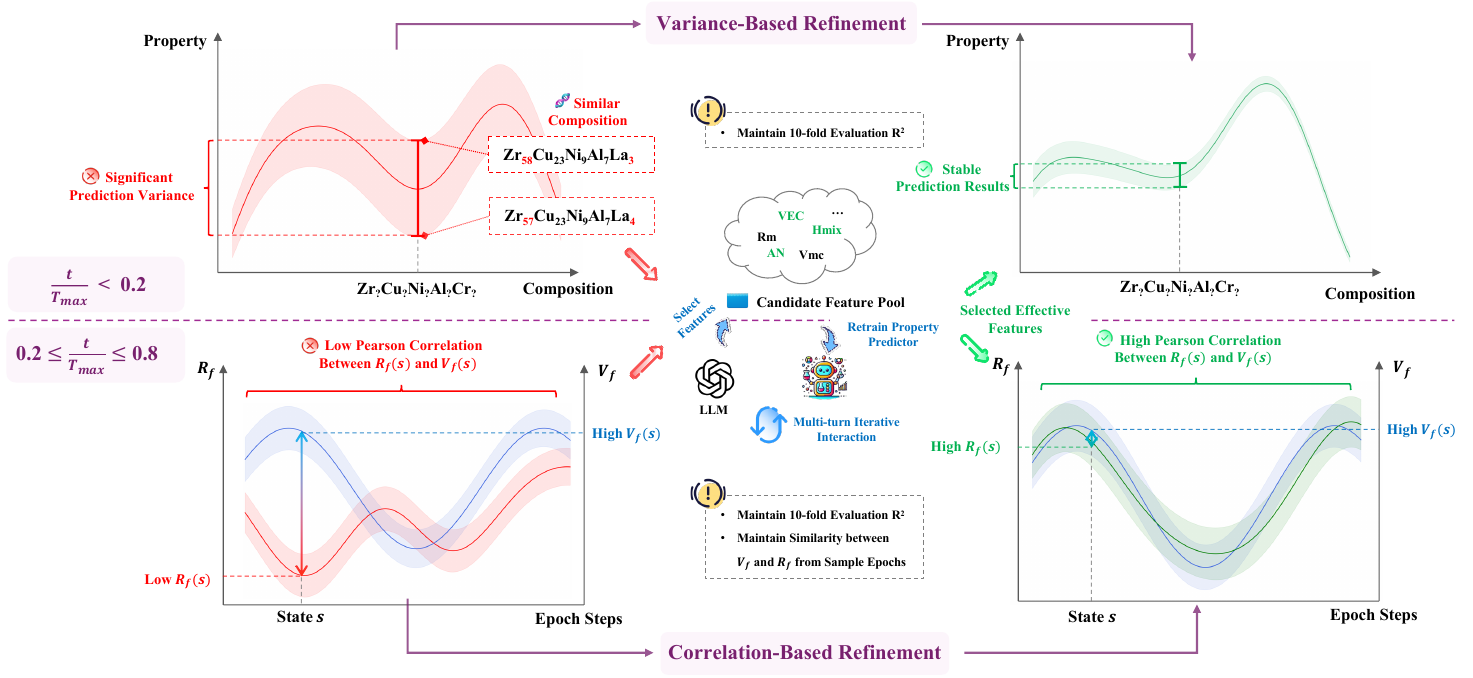}
	\caption{Automatic Model Refinement (AMR) via Adaptive Feedback Loops with Dual-Stage Refinement for Reliable Materials Design.} 
	\label{fig:model_refine}
\end{figure}

To address these issues, we propose an innovative method for \textbf{Automatic Model Refinement (AMR) via Adaptive Feedback Loops}, as shown in Fig.~\ref{fig:model_refine}, which aims to enhance the reliability of materials inverse design by intelligently correcting the guiding model.

\noindent \textbf{Triggering Conditions}: The AMR mechanism includes two key optimization strategies that are triggered at different training stages, ensuring that the model gradually improves and optimizes stably:
\begin{enumerate}
	\item[(1)] \textbf{Variance-Based Refinement} (the upper part of Fig.~\ref{fig:model_refine}):
	In the early stages of RL model training (first 20\% of steps), the state value function ($V_f$) is unstable and cannot provide effective guidance. During this phase, as the RL model explores similar compositions, we assess the prediction variance of the guiding model. If the variance exceeds a threshold, the AMR mechanism is triggered. Optimization is considered effective if:
	\begin{itemize}
		\renewcommand{\labelitemi}{$\circ$}
		\item The average $R^2$ from 10-fold cross-validation should not be lower than the $R^2$ when model were not refined.
		\item The prediction variance should be below the set threshold.
	\end{itemize}
	\item[(2)] \textbf{Correlation-Based Refinement} (the lower part of Fig.~\ref{fig:model_refine}):
	In the mid-stage of training (20\% to 80\% of steps), the RL model's state value function stabilizes. We evaluate the optimization need by calculating the \textit{Pearson Correlation} between the reward curve ($R_f$) and the state value curve ($V_f$). If the correlation coefficient falls below a threshold, the LLM optimization is triggered. Optimization is considered effective if:
	\begin{itemize}
		\renewcommand{\labelitemi}{$\circ$}
		\item The corrected model’s average $R^2$ from 10-fold cross-validation exceeds that without refinement.
		\item The \textit{Pearson Correlation} between $R_f$ and $V_f$ exceeds the threshold.
	\end{itemize}
\end{enumerate}

\noindent \textbf{Refinement Process} (the middle part of Fig.~\ref{fig:model_refine}):
LLM selects 1-3 features from the candidate features based on atomic characteristics and relevant materials knowledge databases~\cite{xiong2020machine2}, which are then added to the guiding model. 
The guiding model is retrained with these expanded features. If the optimization does not meet expectations, the process will be abandoned after a maximum of three iterations.

Overall, the AMR mechanism dynamically adjusts the optimization strategy by leveraging the characteristics of different stages during the RL training process. 
This enables automatic optimization and correction of the guiding model, even in the context of scarce data and insufficient model adaptability. 

The mechanism not only improves the predictive accuracy and robustness of the guiding model but also enhances the framework’s ability to adapt to complex environments and self-correct through the integration of LLMs and materials knowledge databases. 
Additionally, the AMR mechanism ensures consistency between the reward signal and the state value function, promoting stable learning and efficient exploration of the RL model in high-dimensional design spaces.

\begin{algorithm}[h]
	\caption{\textsc{AIMatDesign} Training with Automatic Model Refinement (AMR)}
	\label{alg:AIMatDesign}
	\begin{algorithmic}[1]
	  \Require Dataset $\mathcal{D}$; classifier $f_{\mathrm{cls}}$; regressor $f_{\mathrm{reg}}$;  
			  RL agent $\pi$ with parameters $\theta$; LLM $f_{\mathrm{llm}}$;  
			  max steps $T_{\max}$; steps/epoch $T_{\text{ep}}$; LLM-reward weight $\beta$;  
			  variance threshold $\tau$; correlation threshold $\rho$
	  \Ensure  Optimized RL agent $\pi^{\star}$
	  \State Initialize experience pool $\mathcal{E} \leftarrow \emptyset$ and Build $\mathcal{E}_{tep}$ from $\mathcal{D}$  (difference-based sampling)
	  \State $t \leftarrow 0$  \Comment{global training-step counter}
	
	  \While{$t < T_{\max}$}
		  \For{$k \gets 1$ to $T_{\text{ep}}$}   \Comment{one training epoch}
			  \State $s_t \leftarrow$ current state
			  \State $a_t \leftarrow \pi_\theta(s_t)$
			  \State $s_{t+1} \leftarrow \mathrm{Env_{Step}}(s_t,a_t)$
			  \State $(r_t,\text{done}) \leftarrow f_r(s_t,a_t,s_{t+1})$  \Comment{base reward}
	
			  \If{$t \ge 0.8 \times T_{\max}$ \textbf{and} $f_{\mathrm{cls}}(s_{t+1}) > 0.8$}
				  \State $r_{\mathrm{llm}} \leftarrow f_{\mathrm{llm}}\bigl(s_{t+1},\,f_{\mathrm{reg}}(s_{t+1})\bigr)$
				  \State $r_t \leftarrow (1 - \beta) \times r_t + \beta \times r_{\mathrm{llm}}$   \Comment{Knowledge-Based Reward}
			  \EndIf
	
			  \State $\mathcal{E} \leftarrow \mathcal{E}\cup\{(s_t,a_t,s_{t+1},r_t,\text{done})\}$ \Comment{store experience}
			  \State $t \leftarrow t+1$
			  \If{\textbf{done}} \State \textbf{break} \EndIf
		  \EndFor
		  \State \(\mathcal{B} = \{s_{t'}\}_{t'=t-k+1}^t,\qquad \mathcal{R} = \{r_{t'}\}_{t'=t-k+1}^t,\qquad \mathcal{V} = \{V_{f}(s_{t'})_{\pi}\}_{t'=t-k+1}^t\)
		  \If{$t < 0.2 \times T_{\max}$ \textbf{and} $\operatorname{Var}\bigl(f_{\mathrm{reg}}(\mathcal{B})\bigr) > \tau$}
			  \State \Call{VarianceBasedRefinement}{$\mathcal{B},\,f_{\mathrm{reg}},\,f_{\mathrm{llm}}$} \Comment{Variance-Based Refinement}
		  \ElsIf{$0.2 \times T_{\max} \le t \le 0.8 \times T_{\max}$ \textbf{and} $\operatorname{Corr}(\mathcal{R},\mathcal{V}) < \rho$}
			  \State \Call{CorrelationBasedRefinement}{$\mathcal{B},\,f_{\mathrm{reg}},\,f_{\mathrm{llm}}$} \Comment{Correlation-Based Refinement}
		  \EndIf
		  \State Sample mini-batch $\mathcal{E}_{\text{batch}}$ from $\mathcal{E}$
		  \State Replace part of $\mathcal{E}_{\text{batch}}$ with samples from $\mathcal{E}_{\text{tep}}$ based on $\mathcal{R}$ \Comment{TEP sampling}
		  \State Update $\theta$ with $\mathcal{E}_{\text{batch}}$ using the RL algorithm \Comment{e.g., TD3, PPO, etc.}
	  \EndWhile
	  \State \Return $\pi_\theta$ as $\pi^{\star}$
	\end{algorithmic}
\end{algorithm}

\section{Results}
\label{Results}

The experimental results are divided into three main categories: dataset description (\textsection~\ref{dataset}), ML model modeling results (\textsection~\ref{guidance_model}), RL modeling and exploration outcomes (\textsection~\ref{rl_results}, \textsection~\ref{AMR_results} and \textsection~\ref{design_results}), and ablation experiments (\textsection~\ref{ablation_results}).

These results are based on the implementation details provided in \textsection~\ref{A:implementation_details}, which describe the modeling procedures, and on the KBR and AMR prompt templates in \textsection~\ref{A:prompt_templates}.

\subsection{Experimental Dataset}
\label{dataset}
\begin{table}[ht]
	\centering
	\caption{Statistical Summary of the Experimental Dataset for Amorphous Alloys}
	\label{tab:dataset_desc}
	\resizebox{\textwidth}{!}{%
	\renewcommand{\arraystretch}{1.25} 
	\begin{tabular}{clccccccc}
	\toprule
	\multicolumn{1}{c}{\textbf{Attribute Name}} &
	  \multicolumn{1}{c}{\textbf{Description}} &
	  \textbf{Unit} &
	  \textbf{Count} &
	  \textbf{Mean} &
	  \textbf{Std} &
	  \textbf{Min} &
	  \textbf{80\%} &
	  \textbf{Max} \\ \midrule
	\multicolumn{9}{l}{\textit{\textbf{Regression Dataset}}}                                                  \\
	\textbf{$\bm{D_{max}}$}    & Max diameter           & mm  & 812  & 5.44    & 5.42   & 0.06  & 8      & 35   \\
	\textbf{Tg}      & Glass transition temp. & K   & 878  & 625.91  & 171.62 & 293   & 780    & 1135 \\
	\textbf{Tl}      & Melting temp.          & K   & 820  & 677.04  & 175.98 & 293   & 832    & 1019 \\
	\textbf{Tx}      & Decomposition temp.    & K   & 815  & 1076.76 & 265.55 & 581   & 1309.2 & 1725 \\
	\textbf{$\bm{\sigma_Y}$}   & Yield strength         & MPa & 334  & 1548.69 & 495.05 & 140.5 & 1843   & 4014 \\
	\textbf{$\bm{E}$} & Young's modulus        & GPa & 399  & 94.65   & 52.51  & 16    & 122.8  & 309  \\
	\textbf{$\bm{\varepsilon}$}                & Elongation             & \%  & 296  & 9.98   & 12.52   & 0     & 15  & 75   \\ \midrule
	\multicolumn{9}{l}{\textit{\textbf{Classification Dataset}}}                                              \\
	\textbf{RMG}              & Ribbon Metallic Glass  & -   & 3675 & \multicolumn{5}{c}{\multirow{3}{*}{-}}   \\
	\textbf{CRA}              & Cystalline Alloy       & -   & 1756 & \multicolumn{5}{c}{}                     \\
	\textbf{BMG}              & Bulk Metallic Glass    & -   & 1433 & \multicolumn{5}{c}{}                     \\ \bottomrule
	\end{tabular}%
	}
\end{table}

As shown in Table~\ref{tab:dataset_desc}, the amorphous alloys dataset comprises two subsets: regression and classification.
Material composition features are based on 52 alloy elements, with each sample containing 3-9 valid elements (atomic percentages summing to 100\%), resulting in a sparse distribution in the high-dimensional feature space. 
This poses challenges for both feature learning in machine learning models and exploration strategies in reinforcement learning.

The regression dataset includes three performance categories: (1) \textbf{Geometric properties}: maximum diameter ($D_{\text{max}}$); (2) \textbf{Thermal properties}: glass transition temperature ($T_{\text{g}}$), liquidus temperature ($T_{\text{l}}$), and crystallization temperature ($T_{\text{x}}$); (3) \textbf{Mechanical properties}: yield strength ($\sigma_{\text{y}}$), Young's modulus ($E$), and elongation ($\varepsilon$). 
The sample size for geometric and thermal parameters is approximately $10^3$, while for mechanical properties, it is $10^2$. The dataset includes BMGs and other alloys to improve model generalization.

The classification dataset uses a three-class framework, with ribbon-like metallic glasses (RMG, 3675 samples), crystalline alloys (CRA, 1756 samples), and bulk metallic glasses (BMG, 1433 samples). 
The BMG class represents 21\% of the total, creating an imbalanced distribution. The classification model must handle this imbalance by using probabilistic outputs to quantify the likelihood of a composition being BMG, which aids decision-making in reinforcement learning.

This classification and regression dataset provides essential support for the reinforcement learning environment: classification outputs serve as feasibility constraints, and regression predictions inform the multi-objective reward function, ensuring that generated materials maintain BMG attributes while optimizing overall performance.

\subsection{Guidance Model Development}
\label{guidance_model}

\subsubsection{Classification Modeling}
\label{classification_model}

For the material classification task, we construct a probabilistic output classification model $f_c: \mathbb{R}^{52} \to [0, 1]$, whose output is mapped to the reinforcement learning reward signal $r_c \in [-0.5, 0.5]$ through a linear transformation. To address the 21\% class imbalance, we use the \textit{SMOTE over-sampling technique} to augment BMG samples. Additionally, to ensure model prediction accuracy, we conduct a baseline comparison of several classification models, selecting the best-performing model via 5-fold cross-validation:

\begin{enumerate}
	\item \textbf{Linear Models}: Logistic Regression (LR)~\citep{lavalley2008logistic} and Linear Discriminant Analysis (LDA)~\citep{xanthopoulos2013linear}, which classify based on linear decision boundaries, offering high computational efficiency suitable for initial modeling or simple tasks.
	\item \textbf{Kernel Methods}: Support Vector Classifier (SVC)~\citep{lau2003online}, which uses a kernel function to map data to a higher-dimensional space and excels in tasks with complex decision boundaries.
	\item \textbf{Tree Models}: Decision Tree (DT)~\citep{ying2015decision} and Random Forest (RF)~\citep{breiman2001random}, capable of handling nonlinear features and high-dimensional data, common choices for complex classification tasks.
	\item \textbf{Boosting Methods}: Gradient Boosting Machine (GBM)~\citep{natekin2013gradient}, XGBoost~\citep{chen2016xgboost}, CatBoost~\citep{prokhorenkova2018catboost}, and AdaBoost~\citep{zhu2009multi}, which sequentially optimize the performance of weak classifiers to improve overall prediction capability.
	\item \textbf{Distance-Based Models}: K-Nearest Neighbors (KNN)~\citep{kramer2013k}, which classifies based on sample distances, simple and intuitive but computationally demanding.
	\item \textbf{Probabilistic Models}: Gaussian Naive Bayes (GNB)~\citep{ontivero2017fast}, Multinomial Naive Bayes (MNB)~\citep{abbas2019multinomial}, and Bernoulli Naive Bayes (BNB)~\citep{murphy2006naive}, which classify based on feature probability distributions with strong assumptions about the data.
	\item \textbf{Discriminant Analysis Models}: Quadratic Discriminant Analysis (QDA)~\citep{tharwat2016linear}, which performs well when the data distribution is nonlinear.
\end{enumerate}

\begin{figure}[ht]
	\centering
	\includegraphics[width=.8\textwidth]{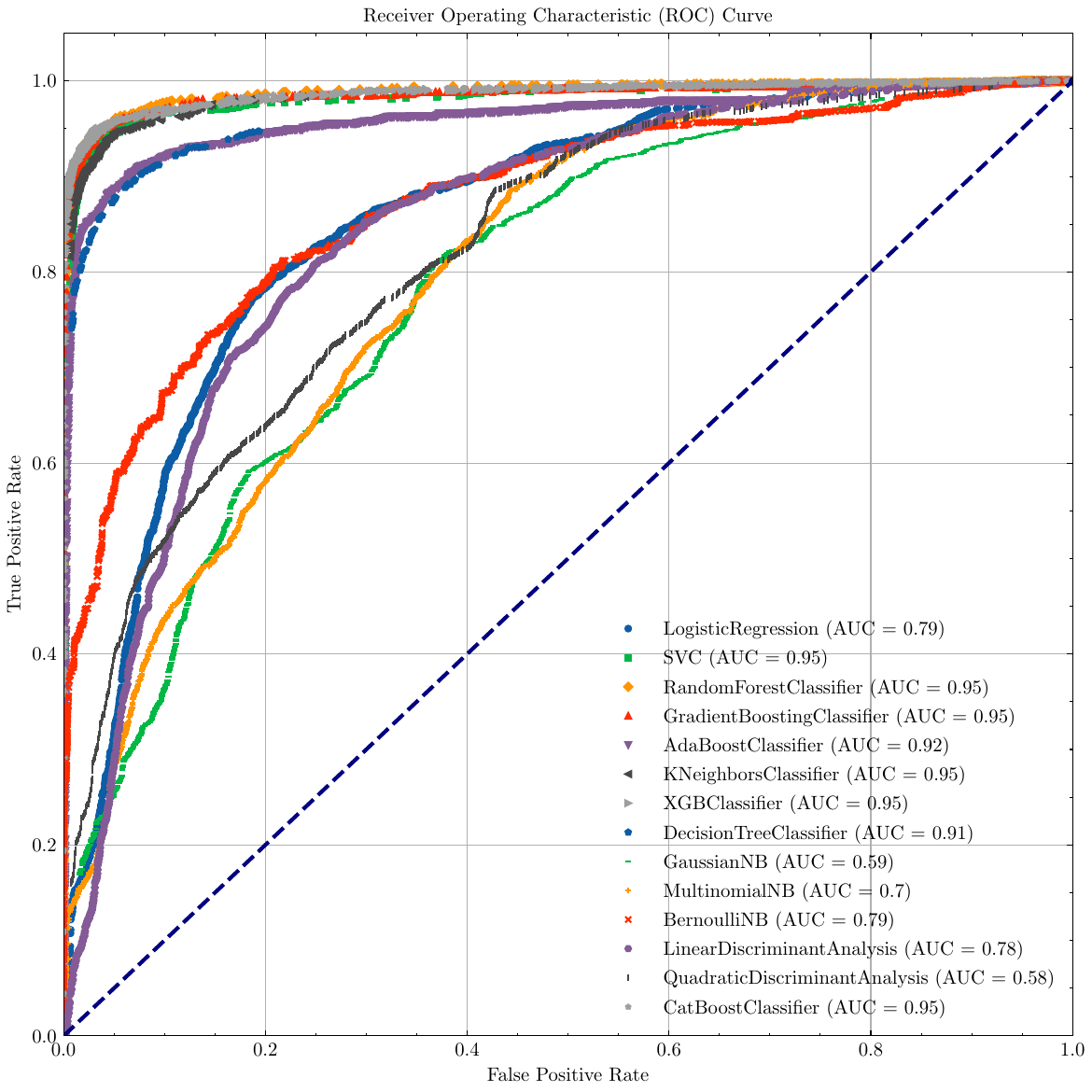}
	\caption{ROC Curve and AUC for Classification Models.}
	\label{fig:cls_results}
\end{figure}

The 5-fold cross-validation performance comparison of classification models is shown in Fig.~\ref{fig:cls_results} (detailed metrics in \textsection \ref{A:cls_results} Table~\ref{tab:cls_results}). Both CatBoost and RF performed best overall. 
However, CatBoost particularly excelled in the BMG classification task with fewer samples, achieving higher Recall scores, which indicates its ability to identify more BMG samples and effectively avoid missing potential high-value targets during RL exploration. 
Therefore, \textbf{we selected CatBoost as the guiding model for the BMG classification task in the virtual environment}. Additionally, CatBoost achieved an AUC of 0.96, demonstrating its strong ability to distinguish between positive and negative samples, providing stable and reliable feedback for RL.

\subsubsection{Regression Modeling}
\label{regression_model}
For the regression task, we construct a multi-task regression model $f_r: \mathbb{R}^{52} \to \mathbb{R}^7$ for quantifying rewards in the range $[0.5, 1]$. 
To ensure prediction accuracy, we conducted a baseline comparison of several regression models and selected the best-performing model using 5-fold cross-validation:
\begin{enumerate}
	\item \textbf{Linear Models}: Ridge regression~\citep{mcdonald2009ridge}, Lasso regression~\citep{ranstam2018lasso}, and ElasticNet~\citep{zou2005regularization}, which fit performance parameters with linear functions and address multicollinearity through regularization.
	\item \textbf{Kernel Methods}: Support Vector Regression (SVR)~\citep{awad2015support}, which uses kernel functions to map the feature space, suitable for high-dimensional sparse data prediction tasks.
	\item \textbf{Tree Models}: Random Forest Regressor (RF)~\citep{breiman2001random}, which captures nonlinear relationships between features through a tree structure, commonly used for complex regression tasks.
	\item \textbf{Boosting Methods}: AdaBoost Regressor~\citep{solomatine2004adaboost}, Gradient Boosting Regressor (GBM)~\citep{natekin2013gradient}, and XGBoost~\citep{chen2016xgboost}, which integrate multiple weak regressors to improve prediction performance, suitable for various task scenarios.
	\item \textbf{Distance-Based Models}: K-Nearest Neighbors Regressor (KNN)~\citep{kramer2013k}, which makes predictions based on neighborhood sample characteristics, suitable for local pattern recognition tasks but less efficient for large-scale data.
	\item \textbf{Randomized Models}:  enhanced deep Random Vector Function Cascade Model (edRVFL)~\citep{hu2022ensemble}, A fast learning model based on random weights, combining recursive and vectorized structures, particularly well-suited for high-dimensional complex regression tasks.
\end{enumerate}

\begin{figure}[htbp]
    \centering
	\begin{subfigure}{0.23\textwidth}
        \includegraphics[width=\linewidth]{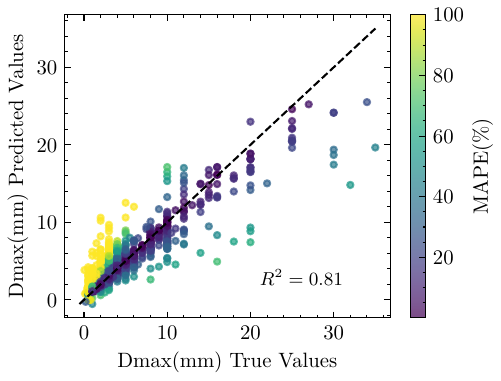}
        \caption{$D_{\text{max}}$(mm)}
    \end{subfigure}
	\hfill
    \begin{subfigure}{0.23\textwidth}
        \includegraphics[width=\linewidth]{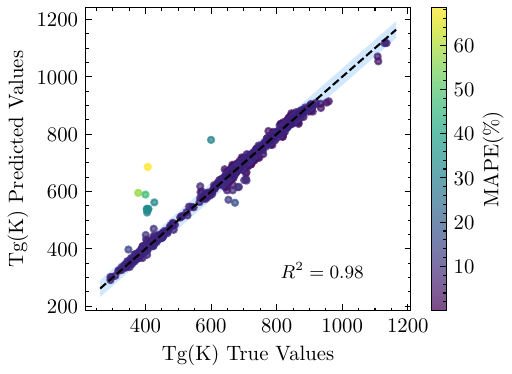}
        \caption{Tg(K)}
    \end{subfigure}
    \hfill
    \begin{subfigure}{0.23\textwidth}
        \includegraphics[width=\linewidth]{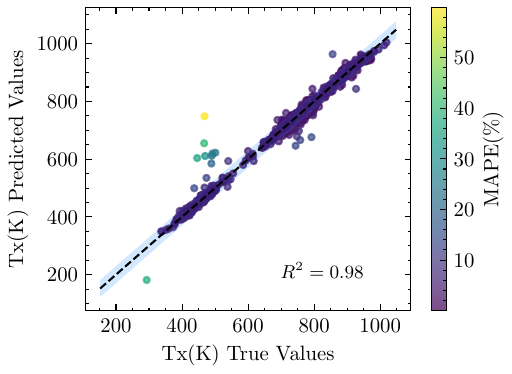}
        \caption{Tx(K)}
    \end{subfigure}
	\hfill
	\begin{subfigure}{0.23\textwidth}
        \includegraphics[width=\linewidth]{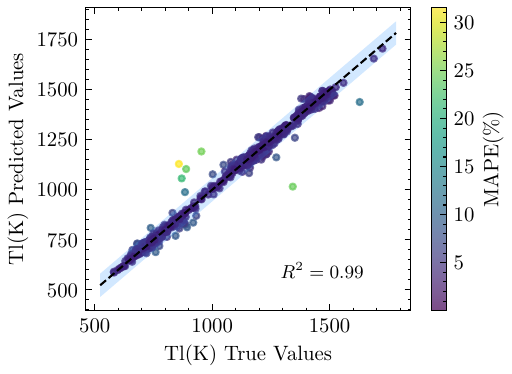}
        \caption{Tl(K)}
    \end{subfigure}

    \hspace{\fill} 
    \begin{subfigure}{0.23\textwidth}
        \includegraphics[width=\linewidth]{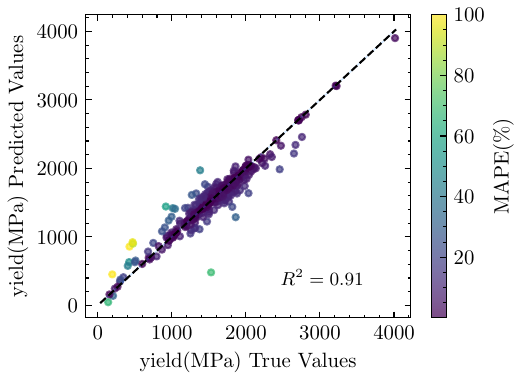}
        \caption{$\sigma_Y$(MPa)}
    \end{subfigure}
    \hfill
    \begin{subfigure}{0.23\textwidth}
        \includegraphics[width=\linewidth]{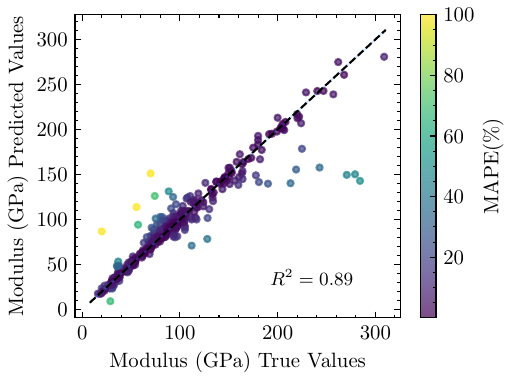}
        \caption{$E$(GPa)}
    \end{subfigure}
	\hfill
    \begin{subfigure}{0.23\textwidth}
        \includegraphics[width=\linewidth]{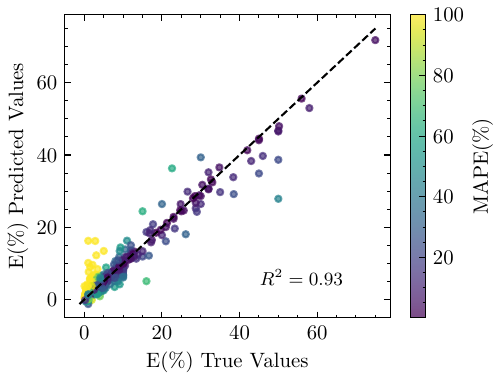}
        \caption{$\varepsilon(\%)$}
    \end{subfigure}
    \hspace{\fill} 

    \caption{Scatter plots of the edRVFL model's regression results, showing the predicted vs. actual values for various material properties.}
	\label{fig:regression_results}
\end{figure}

Fig.~\ref{fig:regression_results} shows the performance of the edRVFL model, which performed best. Details for other models and metrics are in \textsection~\ref{A:regression_results} Table~\ref{tab:regression_results}.
When predicting geometric properties, thermal properties, and mechanical properties of material compositions, edRVFL outperformed all other models across key metrics ($R^2$, RMSE, and MAPE). 
Notably, edRVFL improved $R^2$ by over 0.27 for predicting $\varepsilon (\%)$ and over 0.3 for $\sigma_{\text{Y}}$ (MPa). 
Additionally, edRVFL demonstrated stable and high-precision performance in predicting other properties. Therefore, \textbf{we selected edRVFL as the guiding model for performance prediction in the virtual environment.} 

The regression model provides quantifiable feedback for the reward function in the virtual environment, with edRVFL’s high $R^2$ ensuring accurate material property predictions and significantly reducing strategy bias caused by prediction errors, thus enhancing the RL model's exploration efficiency and reliability in new material design.

\subsubsection{Trustworthy Experience Pool's Distribution}
\label{TEP_distribution}

\begin{figure}[ht]
	\centering
	\includegraphics[width=.7\textwidth]{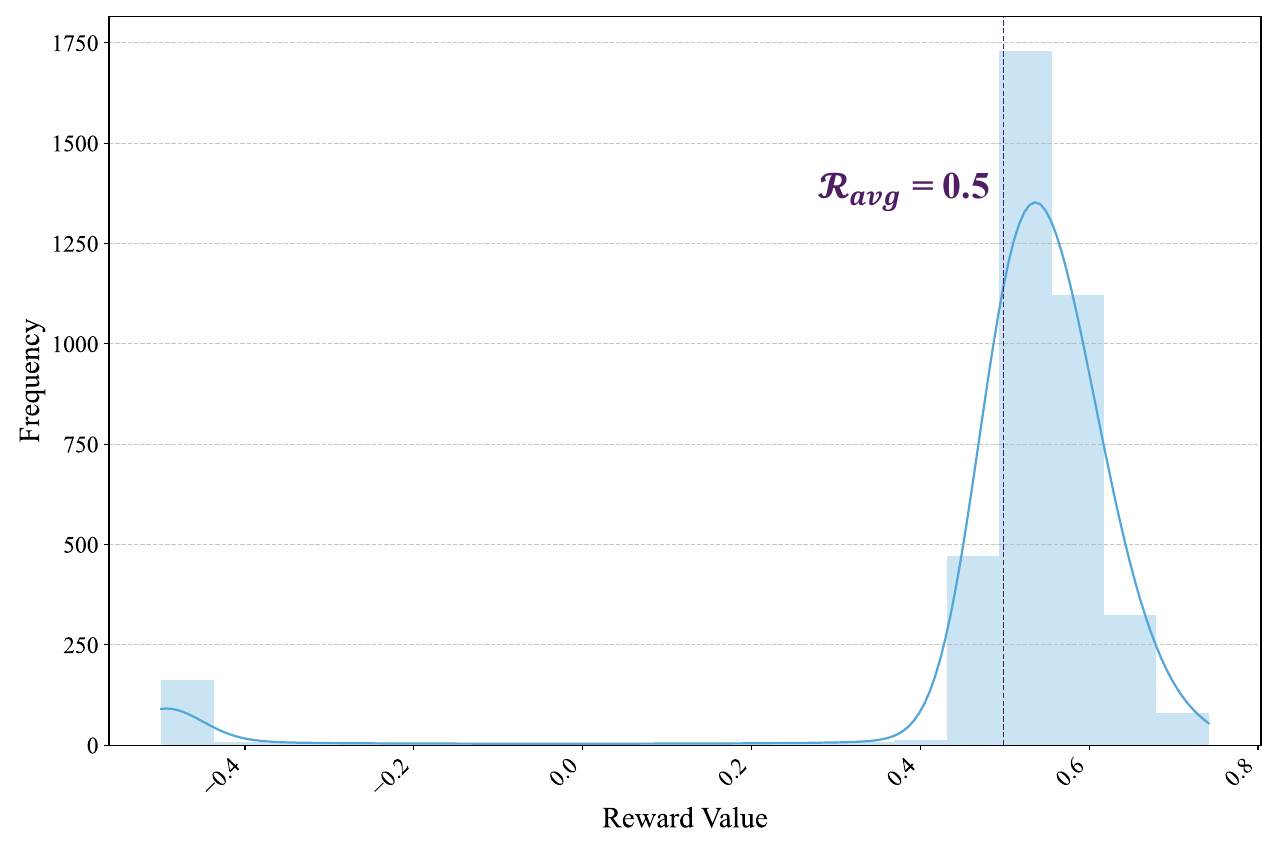}
	\caption{Distribution of Reward Values in the Trustworthy Experience Pool.}
	\label{fig:reward_distribution}
\end{figure}

As shown in Fig.~\ref{fig:reward_distribution}, the reward values in the experience pool exhibit a unimodal distribution, with over 95\% of the samples concentrated in the range $[0.4, 0.6]$, and a mean of $0.5$. 
This distribution indicates that most experience samples provide positive feedback for policy optimization. 
At the same time, experiences with lower rewards (e.g., in the range $[-0.5, 0]$) correspond to states where $s_2$'s alloy is non-BMG, representing infeasible solutions discovered during exploration, which provide negative feedback constraints for strategy optimization.

As detailed in Methods (\textsection\ref{trustworthy_experience_pool}), we apply a reward-aware replacement strategy that swaps part of each training batch with higher-reward samples from the TEP.  
This simple adjustment accelerates exploration and leads to faster, more stable convergence in subsequent epochs.

\subsection{RL Design Results}
\label{rl_results}

\noindent \textbf{Experimental Setup} We first analyzed the dataset from \textsection~\ref{dataset} and identified 35 exploration bases, corresponding to the most prevalent elements in the compositions. 
Based on the compositional ranges provided by the dataset, we set component limits for each base and randomly generated an initial base within this range as the starting state $S_0$ for the RL process. 

During RL training, each epoch consists of up to 128 steps (terminated early if the stopping condition is met), with a total of 1000 epochs. 
Therefore, the theoretical compositional space explored by the RL method is $128 \times 1000$. 
For fairness, the number of ML predictions used by traditional optimization algorithms during the search process is also set to $128 \times 1000$.

Regarding the AMR and KBR components, the LLMs used were \emph{GPT-4o-2025-03-26}~\citep{hurst2024gpt}.

\noindent \textbf{Baselines} To validate the effectiveness of reinforcement learning in materials design tasks, we compared multiple traditional optimization methods and RL algorithms, categorizing the baseline methods into three types:
\begin{enumerate}
	\item \textbf{Traditional Inverse Design Algorithms}: These rely on search and evolutionary strategies, using heuristic rules for material optimization. 
	Methods include grid search~\citep{liashchynskyi2019grid}, which uniformly samples different component combinations within a predefined search space, 
	and NSGA-II~\citep{deb2002fast}, a multi-objective optimization method based on genetic algorithms that optimizes material compositions via selection, crossover, and mutation.
	\item \textbf{Value-Based RL}: These use the Q-value function to estimate the optimal policy and perform material selection based on value evaluation. 
	Methods include DQN~\citep{mnih2015human}, which approximates the Q function using deep neural networks and explores using an $\epsilon$-greedy strategy.
	\item \textbf{Policy-Based RL}: These directly optimize the policy network to enable the model to autonomously generate material compositions. 
	Methods include DDPG~\citep{lillicrap2015continuous}, which optimizes the policy in continuous action spaces via the Actor-Critic mechanism; 
	TD3~\citep{fujimoto2018addressing}, which introduces twin Q networks and delayed updates to improve stability; 
	SAC~\citep{haarnoja2018soft}, which incorporates entropy regularization to enhance exploration and mitigate overfitting; 
	and PPO~\citep{schulman2017proximal}, which employs trust region optimization to constrain policy updates for improved training efficiency and stability.
\end{enumerate}

\noindent \textbf{Evaluation Metrics} To comprehensively evaluate the performance of each method in materials design, the following key metrics were used in the experiments:
\begin{itemize}
	\renewcommand{\labelitemi}{$\circ$}
	\item \textbf{$\text{SR}_{\text{legal}}$}: The \textit{step-level} success rate of generating samples that satisfy material design legality constraints. (Since traditional design methods have predefined component ranges, $\text{SR}_{\text{legal}}$ is not reported for these methods.)
	\item \textbf{$\text{SR}_{\text{cls}}$}: The \textit{step-level} classification success rate of generating samples belonging to the target material class (e.g., BMG).
	\item \textbf{$\text{SR}_{\text{80\%}}$}: The \textit{step-level} success rate of generating samples that meet the top 80\% of key performance indicators in the original dataset, including maximum diameter ($D{\text{max}}$), glass transition temperature ratio ($T{\text{g}}/T_{\text{l}}$), yield strength ($\sigma_Y$), Young's modulus ($E$), and elongation ($\varepsilon(\%)$).
	\item \textbf{$\text{SR}_{\text{done}}$}: The \textit{epoch-level} success rate of generating samples that simultaneously meet all design objectives by the end of each training epoch.
\end{itemize}

\begin{table}[ht]
	\centering
	\caption{Comparison of Design Success Rates Across Different Performance Metrics for Traditional and Reinforcement Learning-Based Material Design Methods}
	\label{tab:rl_results}
	\resizebox{\textwidth}{!}{%
	\renewcommand{\arraystretch}{1.3} 
	\begin{tabular}{clcccccccc}
	\toprule
	\multicolumn{1}{l}{\multirow{2}{*}{\textbf{}}} & \multirow{2}{*}{\textbf{Methods}} & \multirow{2}{*}{\textbf{$\text{SR}_{\text{legal}}$}} & \multirow{2}{*}{\textbf{$\text{SR}_{\text{cls}}$}} & \multicolumn{5}{c}{\textbf{$\text{SR}_{\text{80\%}}$}} & \multirow{2}{*}{\textbf{$\text{SR}_{\text{done}}$}} \\ \cline{5-9}
	\multicolumn{1}{l}{} &  &  &  & \textbf{$\bm{D}_{\text{max}}$(mm)} & \textbf{Tg/Tl} & \textbf{$\bm{\sigma_Y}$(MPa)} & \textbf{$\bm{E}$(GPa)} & \textbf{$\bm{\varepsilon(\%)}$} &  \\ \midrule
	\multirow{2}{*}{\textbf{Traditional}} & \textbf{Grid Search}~\citep{liashchynskyi2019grid} & - & 91.37 & 28.64 & 44.73 & 17.61 & 26.41 & 12.77 & 5.83 \\
		& \textbf{NSGA-II}~\citep{deb2002fast} & - & 94.42 & 39.48 & 55.82 & 21.42 & 42.96 & 23.44 & 14.71 \\ \midrule
	\multirow{7}{*}{\textbf{RL}} & \textbf{Random}~\citep{liashchynskyi2019grid} & 92.60 & 90.73 & 31.61 & 49.80 & 16.33 & 35.58 & 16.33 & 7.65 \\
		& \textbf{DQN}~\citep{mnih2015human} & 97.35 & 96.28 & 43.32 & 58.94 & 38.98 & 48.87 & 35.62 & 38.59 \\
		& \textbf{DDPG}~\citep{lillicrap2015continuous} & 98.34 & 98.73 & 48.48 & 62.38 & \underline{40.56} & \underline{52.65} & 41.27 & 43.21 \\
		& \textbf{TD3}~\citep{fujimoto2018addressing} & \underline{99.50} & \textbf{99.37} & 47.63 & 63.40 & 39.98 & 51.23 & \underline{43.43} & \underline{45.32} \\
		& \textbf{SAC}~\citep{haarnoja2018soft} & 97.62 & 98.32 & 46.82 & 57.32 & 36.84 & 48.46 & 36.85 & 40.87 \\
		& \textbf{PPO}~\citep{schulman2017proximal} & 99.36 & 98.69 & \underline{48.56} & \textbf{64.82} & 38.54 & 50.83 & 42.73 & 41.89 \\ \cline{2-10}
		& \textbf{\textsc{AIMatDesign}} & \textbf{99.65} & \underline{99.12} & \textbf{50.94} & \underline{63.58} & \textbf{46.93} & \textbf{55.21} & \textbf{49.38} & \textbf{50.32} \\ \bottomrule
	\end{tabular}%
	}
\end{table}

The experimental results, shown in Table~\ref{tab:rl_results}, demonstrate that our model exhibits significant advantages in multi-objective materials inverse design, achieving near-theoretical limits in both legality constraint success rate ($\text{SR}{\text{legal}}$ = 99.65\%) and material classification success rate ($\text{SR}{\text{cls}}$ = 99.12\%), validating its precise control over complex compositional constraints.

For the key performance indicator ($\text{SR}_{\text{80\%}}$), the model shows improvements of over 6 percentage points compared to the optimal RL baseline in yield strength ($\sigma_Y$(MPa) = 46.93\%) and elongation ($\varepsilon(\%)$ = 49.38\%). 
Additionally, the overall success rate ($\text{SR}_{\text{done}}$ = 50.32\%) is 3.4 times higher than that of the traditional evolutionary algorithm NSGA-II, highlighting the efficient exploration capabilities of reinforcement learning in continuous high-dimensional spaces.

It is noteworthy that, under the same number of ML predictions (128,000), traditional methods suffer from a low proportion of valid samples (less than 15\%) due to their random search nature. 
In contrast, our model achieves goal-directed compositional generation through a dynamic policy network, providing a more efficient solution for high-cost material experiments.

\begin{figure}[htbp]
	\centering
	\includegraphics[width=\textwidth]{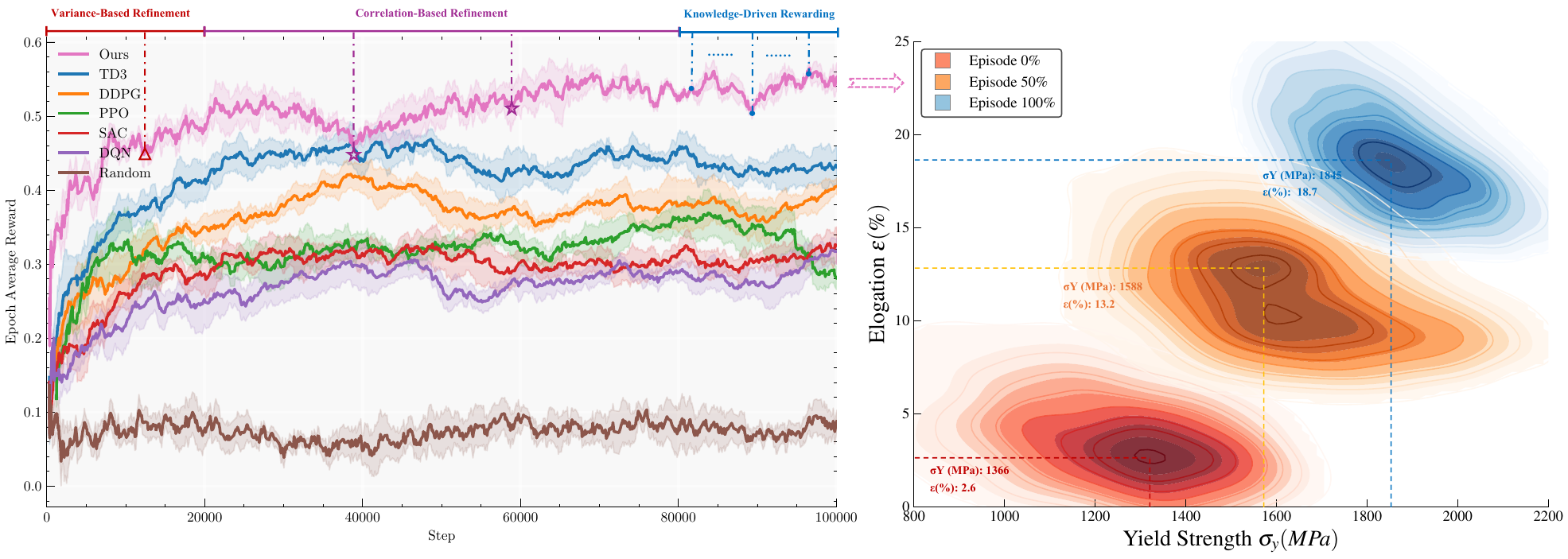}
	\caption{Comparison of Training Episode Results: Left - Reward Progression of RL-Based Models; Right - Performance Evolution of the \textsc{AIMatDesign} Model}
	\label{fig:train_episode_results}
\end{figure}

In the left half of Fig.~\ref{fig:train_episode_results}, the model demonstrates a significant improvement in convergence speed through the TEP, with an average reward increase of 0.1 in the first 5000 training steps compared to the TD3 algorithm. 
During training, two optimization mechanisms are triggered sequentially: \textbf{Variance-Based Refinement} at episode 201, and \textbf{Correlation-Based Refinement} at episodes 398 and 503. 
The experimental results show that without model optimization, the reward metric declines due to the performance limitations of the initial machine learning guiding model (e.g., a decrease of 0.05 at episode 398). 
However, after optimization, the model performance improves significantly. In the later training stages (last 20\% of steps), the introduction of the KBR facilitates secondary optimization of the converged model, leading to a 0.05 increase in the reward curve.

The right half of Fig.~\ref{fig:train_episode_results} shows that as \textsc{AIMatDesign} Model training progresses, the distribution of the generated BMGs materials' $E$(GPa) and $\sigma_Y$(MPa) performance continuously shifts towards the upper-right region of the coordinate system. The average elastic modulus ($E$) increases by 18.7\%, forming a clear trend of performance improvement.

\subsection{Automatic Model Refinement Results}
\label{AMR_results}

\noindent \textbf{Experimental Setup} The refinement strategies were supported by \emph{GPT-4o-2025-03-26}~\citep{hurst2024gpt}, which interacted with the model using the current predictions and a material knowledge base to select 1-3 features from the candidate pool~\cite{xiong2020machine2}. If optimization failed to meet expectations, up to three iterations were performed before abandoning the attempt.

\begin{figure}[ht]
	\centering
	\includegraphics[width=\textwidth]{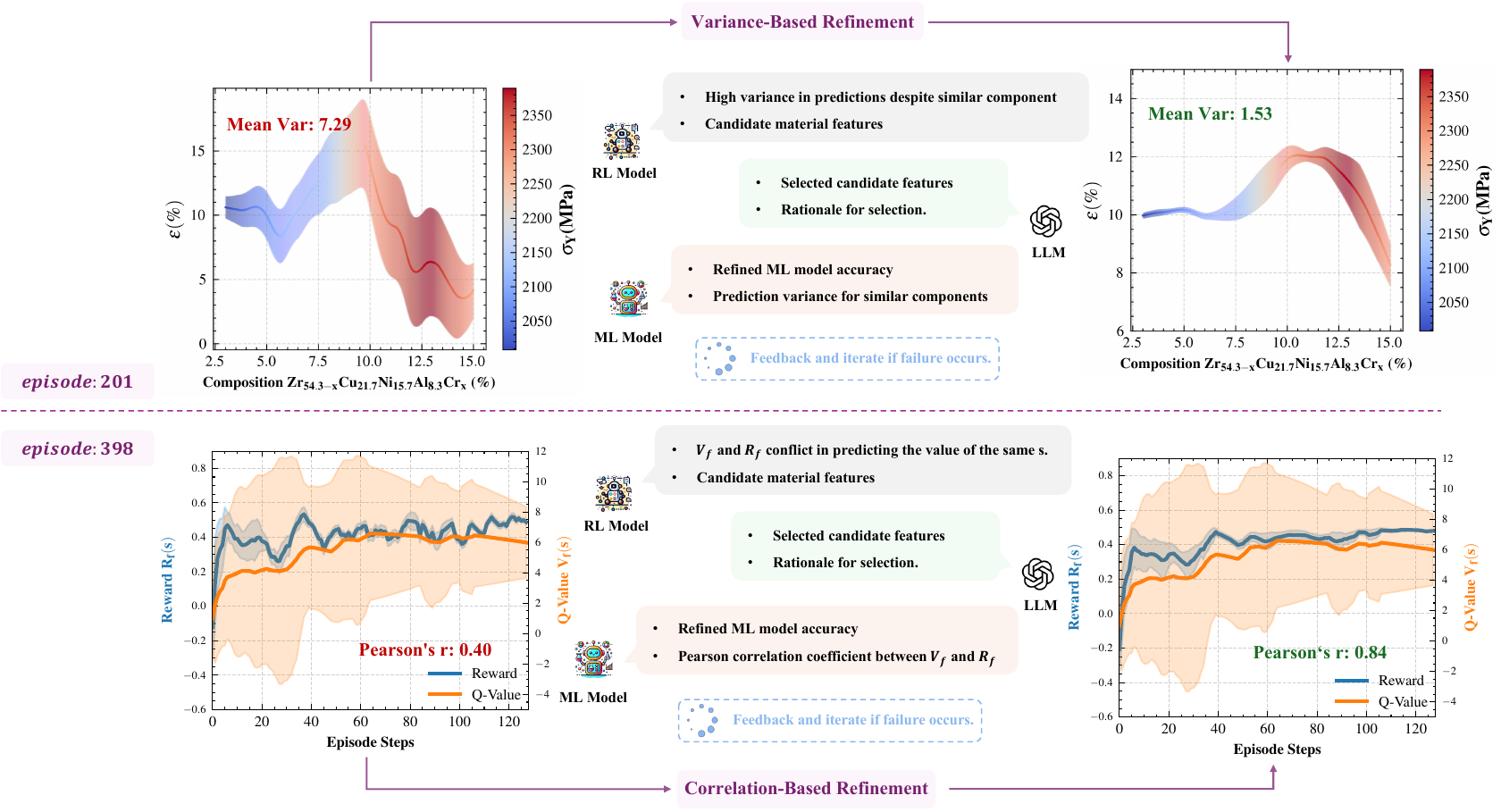}
	\caption{Effectiveness of Automatic Model Refinement illustrated by two cases (not cherry-picked): (a) Variance-Based Refinement reduces local prediction variance for elongation ($\varepsilon$); (b) Correlation-Based Refinement enhances consistency between the ML model's reward and the RL model's Q-value.}
	\label{fig:refinement_results}
\end{figure}

Fig.~\ref{fig:refinement_results} shows the experimental results obtained using the Variance-Based and Correlation-Based refinement strategies in \textbf{\textsc{AIMatDesign}} training.
\begin{itemize}
	\item \textbf{Variance-Based Refinement}:
	In the $201^{st}$ iteration, the model's prediction of elongation ($\varepsilon$) showed high variance (mean squared error of $7.29$). Through iterative interaction with LLMs, a set of material features was selected from the candidate feature pool and the ML model was retrained, effectively reducing the variance in this region to $1.53$, thus minimizing the potential uncertainty caused by high variance.
	\item \textbf{Correlation-Based Refinement:}
	In the $398^{th}$ iteration, the correlation between the reward curve provided by the ML model and the Q-value curve predicted by the RL model was low (Pearson correlation coefficient of only $0.40$). After LLMs' interactive analysis and selecting applicable material features, the correlation coefficient was successfully increased to $0.84$. This not only ensured consistency between the two models but also significantly reduced the fluctuations in the reward and Q-value, thereby enhancing the overall decision stability.
\end{itemize}

Overall, \textbf{Variance-Based Refinement} targets regions with high local variance, optimizing prediction accuracy at a fine-grained level, while \textbf{Correlation-Based Refinement} aims to improve the correlation between global performance metrics to enhance decision consistency between the RL and ML models. 
Together, these strategies complement each other, providing strong support for efficient exploration and reliable decision-making in RL-based new materials design.

\subsection{Ablation Study}
\label{ablation_results}

Table~\ref{tab:ablation_results} compares the performance of the full model with models where certain components (Trustworthy Experience Pool, Automatic Model Refinement, and Knowledge-Based Reward) are removed. 
The results show that removing any component leads to a decline in overall performance, while the full model performs best across several key metrics, confirming the positive contribution of each component to the overall framework.

\begin{table}[ht]
	\centering
	\caption{Ablation Study of Key Components in the \textsc{AIMatDesign} Framework}
	\label{tab:ablation_results}
	\resizebox{.92\textwidth}{!}{%
	\renewcommand{\arraystretch}{1.25} 
	\begin{tabular}{lcccccccc}
	\toprule
	\multicolumn{1}{l}{\multirow{2}{*}{\textbf{}}} & \multirow{2}{*}{\textbf{$\text{SR}_{\text{legal}}$}} & \multirow{2}{*}{\textbf{$\text{SR}_{\text{cls}}$}} & \multicolumn{5}{c}{\textbf{$\text{SR}_{\text{80\%}}$}} & \multirow{2}{*}{\textbf{$\text{SR}_{\text{done}}$}} \\ \cline{4-8}
	\multicolumn{1}{l}{} &  &  & \textbf{$\bm{D}_{\text{max}}$(mm)} & \textbf{Tg/Tl} & \textbf{$\bm{\sigma_Y}$(MPa)} & \textbf{$\bm{E}$(GPa)} & \textbf{$\bm{\varepsilon(\%)}$} &  \\ \midrule
	\textbf{TD3} & 99.50 & 99.37 & 47.63 & 63.40 & 39.98 & 51.23 & 43.43 & 45.32 \\
	\textbf{w/o TEP} & 99.35 & 99.23 & \underline{49.32} & \underline{63.82} & \underline{43.56} & 54.35 & 46.87 & 47.63 \\
	\textbf{w/o AMR} & 99.50 & \underline{99.42} & 47.23 & 62.70 & 41.38 & 52.32 & 42.78 & 45.84 \\
	\textbf{w/o KBR} & \underline{99.60} & \textbf{99.48} & 48.74 & \textbf{64.83} & 42.76 & \underline{54.76} & \underline{48.65} & \underline{49.32} \\ \hdashline
	\textsc{AIMatDesign} & \textbf{99.65} & 99.12 & \textbf{50.94} & 63.58 & \textbf{46.93} & \textbf{55.21} & \textbf{49.38} & \textbf{50.32} \\ \bottomrule
	\end{tabular}%
	}
	\end{table}

Specifically, the ``w/o AMR" model shows a 4.5\% decrease in $\text{SR}_{\text{done}}$, indicating that the automatic model refinement process provides an effective feedback mechanism for both the ML and RL models, significantly impacting the final material design success rate. 

Additionally, because the introduction of Correlation-Based Refinement occurs at a fixed point in time, the convergence speed of RL is crucial for subsequent model refinement and design capability. 
Removing the Trustworthy Experience Pool (``w/o TEP") slows down early-stage RL convergence, making it more difficult to fully leverage later refinement, resulting in a lower success rate compared to the full model. 
On the other hand, ``w/o KBR" performs well on local prediction tasks but lags behind the full model in overall success rate ($\text{SR}_{\text{done}}$).

In summary, the full model achieves more balanced and superior performance across all metrics, demonstrating the critical importance of the synergistic effect of the three components for multi-objective optimization and reliable decision-making in RL-based new materials design.

\subsection{Design Results}
\label{design_results}

\begin{figure}
	\centering
	\includegraphics[width=\textwidth]{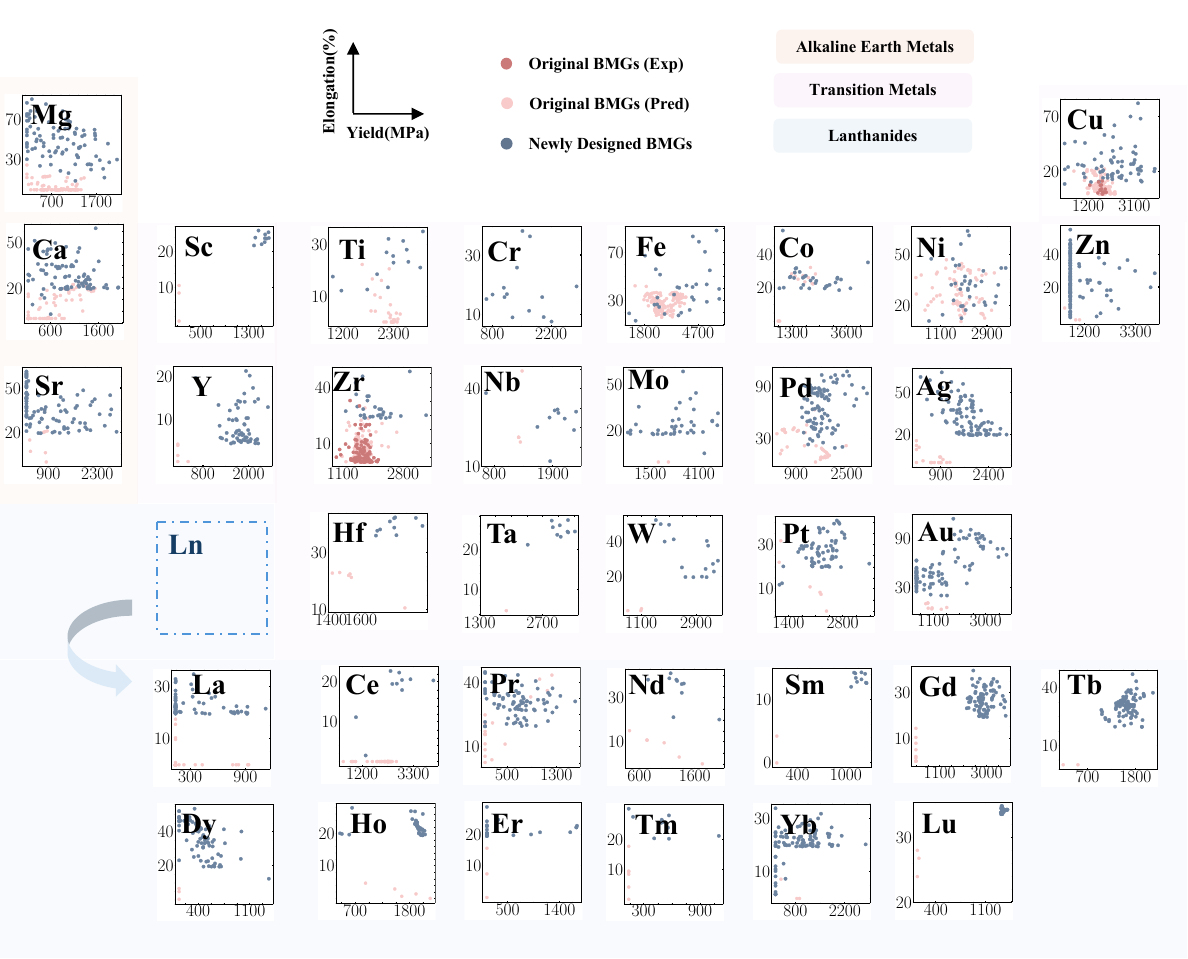}
	\caption{Distribution of $\sigma_Y$(MPa) performance (x-axis) and $\varepsilon(\%)$ (y-axis) across 35 representative metal bases during the design exploration process.}
	\label{fig:design_results}
\end{figure}

To validate the applicability of the proposed method across different base materials, we conducted 100 training epochs on 35 representative metal bases and recorded the $\text{SR}_{\text{done}}$ for each base. 
The results are displayed in the Fig.~\ref{fig:design_results}, with alkaline earth metals (orange), transition metals (purple), and lanthanide elements (blue) showing the distribution of target performance during the exploration process.

The results, shown in Fig.~\ref{fig:design_results}, highlight substantial differences in design difficulty across elements: base elements such as Au, Zn, and Ag achieve an $\text{SR}_{\text{done}}$ of 100\% or close to it, whereas bases like Sm and Ta exhibit markedly lower $\text{SR}_{\text{done}}$ values.
This difference is partly due to the inherent chemical properties and feasible space variations of each element, and also reflects the reinforcement learning strategy's adaptability, which is still constrained by initial conditions and design constraints.

In multiple experiments, the overall average success rate of the method was 54.8\%. 
Notably, even for bases with lower success rates (e.g., Hf, Nb), feasible solutions were still found within a smaller range. 
This demonstrates that the proposed method can provide stable, high success rates for materials with ``easy-to-explore" design spaces, such as noble metals and alkaline earth metals, while also possessing the ability to uncover potential feasible solutions in more challenging material bases (e.g., rare earth or transition metals). 
This approach \textbf{balances broad search capabilities with deep exploration}, offering valuable insights for future RL-based material design iterations.

\begin{figure}
	\centering
	\includegraphics[width=\textwidth]{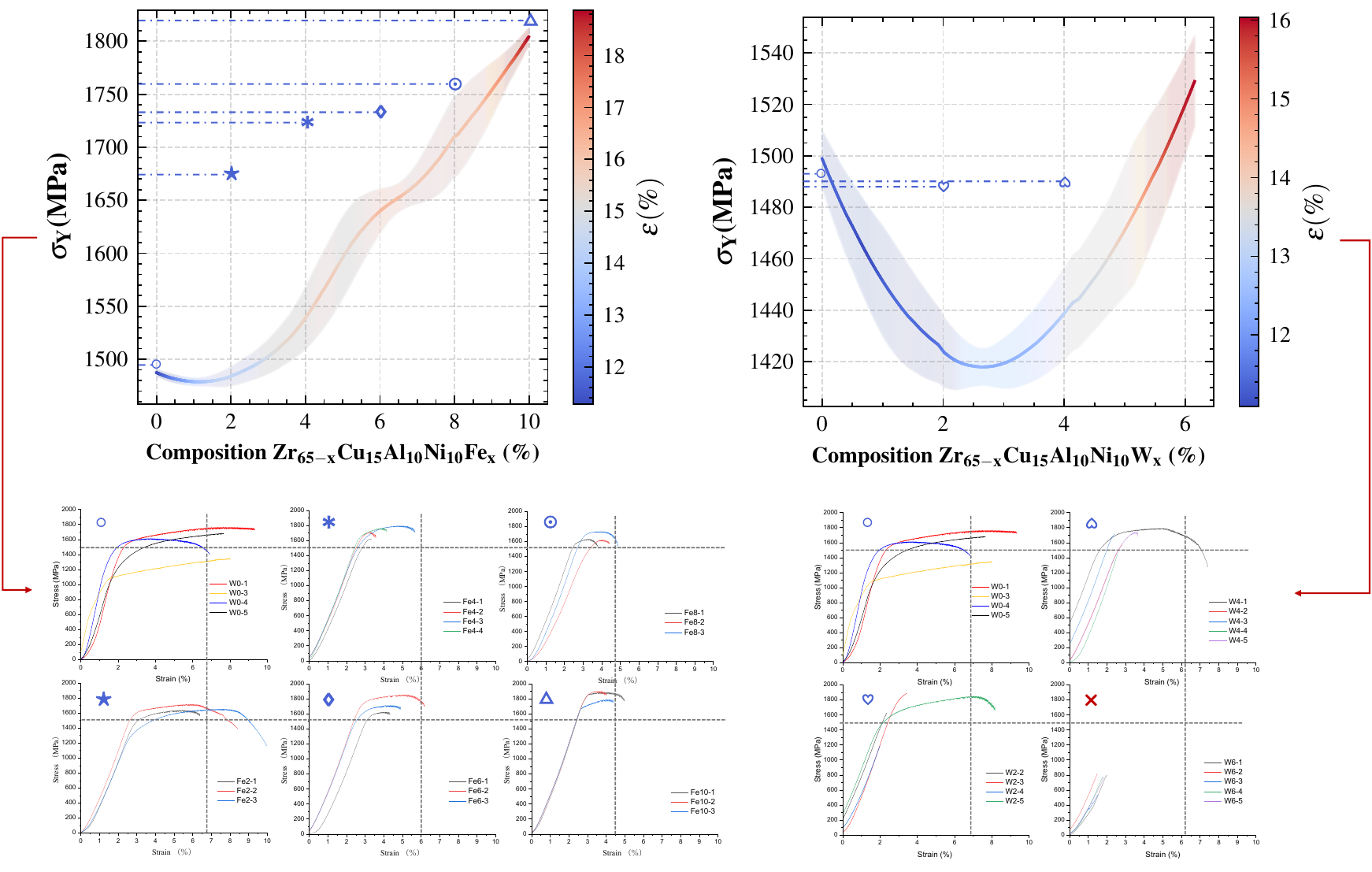}
	\caption{Predicted and experimental mechanical properties of Fe- and W-alloyed Zr-based BMGs. Top: Predicted yield strength ($\sigma_Y$) and plastic strain ($\varepsilon$) trends for Zr$_{65-x}$Cu$_{15}$Al$_{10}$Ni$_{10}$Fe$_x$ (left) and Zr$_{65-x}$Cu$_{15}$Al$_{10}$Ni$_{10}$W$_x$ (right). Bottom: Experimental stress–strain curves. Fe alloying raises $\sigma_Y$ monotonically while $\varepsilon$ peaks at $x=2$ (10.2\%). In contrast, W alloying yields minor strength fluctuations at $x=2\sim4$ but a pronounced strength drop and near‑zero ductility at $x=6$, implying partial crystallisation.}
\label{fig:exp_results}
\end{figure}

For a more rigorous assessment of \textsc{AIMatDesign}, we selected the two Zr‑based BMG cluster centres obtained by $k$‑means~\citep{hartigan1979algorithm} in \textsection~\ref{rl_results}—Zr$_{63}$Cu$_{15}$Al$_{10}$Ni$_{10}$Fe$_{2}$ and Zr$_{63}$Cu$_{15}$Al$_{10}$Ni$_{10}$W$_{2}$—together with their neighbouring compositions (top panel in Fig.\ref{fig:exp_results}), for experimental validation (bottom panel in Fig.~\ref{fig:exp_results}).
The Zr system was chosen because it accounts for the largest share of the original database, yielding the highest model confidence.

All specimens were produced by single‑step suction casting without post‑heat treatment; room‑temperature compression tests were performed at a strain rate of $10^{-4},\mathrm{s}^{-1}$ (Table~\ref{tab:exp_results}). 
The average relative error between predicted and experimental yield strength, $\sigma_Y$, is only 4.9\%, and Fig.~\ref{fig:exp_results} confirms that the experimental $\sigma_Y$ trend mirrors the \textsc{AIMatDesign} prediction.

\begin{table}[ht]
	\centering
	\caption{Experimental validation of \textsc{AIMatDesign} predictions for Zr‑based bulk metallic glasses.}
	\label{tab:exp_results}
	\resizebox{.92\textwidth}{!}{%
	\renewcommand{\arraystretch}{1.05} 
	\begin{tabular}{cccccc}
	\toprule
	& \textbf{Composition} & \textbf{$\bm{\sigma_Y}$ (MPa) $\bm{_{\text{Pred.}}}$} & \textbf{$\bm{\sigma_Y}$ (MPa) $\bm{_{\text{Exp.}}}$} & \textbf{$\bm{\varepsilon(\%)}$ $\bm{_{\text{Pred.}}}$} & \textbf{$\bm{\varepsilon(\%)}$ $\bm{_{\text{Exp.}}}$} \\
	\midrule
	$\circ$ & Zr$_{65}$Cu$_{15}$Al$_{10}$Ni$_{10}$ & 1486 & 1493 & 11.2 & 6.83 \\
	$\bigstar$ & Zr$_{63}$Cu$_{15}$Al$_{10}$Ni$_{10}$Fe$_{2}$ & 1485 & 1671 & 14.3 & 10.2 \\
	$\ast$ & Zr$_{61}$Cu$_{15}$Al$_{10}$Ni$_{10}$Fe$_{4}$ & 1535 & 1722 & 15.5 & 5.8 \\
	$\diamond$ & Zr$_{59}$Cu$_{15}$Al$_{10}$Ni$_{10}$Fe$_{6}$ & 1647 & 1731 & 16.3 & 6.0 \\
	$\odot$ & Zr$_{57}$Cu$_{15}$Al$_{10}$Ni$_{10}$Fe$_{8}$ & 1713 & 1760 & 15.9 & 5.0 \\
	$\triangle$ & Zr$_{55}$Cu$_{15}$Al$_{10}$Ni$_{10}$Fe$_{10}$ & 1789 & 1820 & 18.8 & 4.6 \\ 
	\hdashline
	$\heartsuit$ & Zr$_{63}$Cu$_{15}$Al$_{10}$Ni$_{10}$W$_{2}$ & 1424 & 1488 & 11.7 & 7.8\\
	$\spadesuit$ & Zr$_{61}$Cu$_{15}$Al$_{10}$Ni$_{10}$W$_{4}$ & 1442 & 1490 & 13.0 & 7.0 \\
	\bottomrule
	\end{tabular}
	}
\end{table}

By contrast, the measured plastic strain $\varepsilon$ is systematically lower than predicted owing to two factors:
\begin{enumerate}
	\item[\textit{(i)}] Most training data were taken from literature values for mechanically polished, diameter‑optimised cylindrical samples, whereas the present one‑step cast plates exhibit surface defects and residual stresses that were not explicitly modelled.
	\item[\textit{(ii)}] Process parameters are often missing from the source literature, preventing the model from capturing processing–microstructure–ductility couplings.
\end{enumerate}

Even so, the Zr$_{63}$Cu$_{15}$Al$_{10}$Ni$_{10}$Fe$_{2}$ sample achieved an experimental $\varepsilon$ of 10.2\%, demonstrating that \textsc{AIMatDesign} can deliver BMGs whose yield strength agrees closely with predictions while retaining appreciable ductility without further processing. 
This closed‑loop validation underscores the engineering feasibility of the framework and establishes a paradigm for subsequent iterations on more challenging base alloys.

\section{Conclusion}
\label{conclusion}

This study addresses the challenges of data scarcity and model reliability in exploring high-dimensional materials composition spaces by proposing the RL-based \textsc{AIMatDesign} inverse design framework. 
The method accelerates model convergence through a difference-based augmented trustworthy experience pool and incorporates materials domain expert knowledge at key stages, effectively overcoming the limitations of purely data-driven approaches. 
Additionally, an automated dynamic model refinement strategy is introduced, which not only enhances the stability and convergence efficiency of RL in high-dimensional, complex performance spaces but also provides a more flexible and scalable solution for materials inverse design. 

Experimental results show that \textsc{AIMatDesign} outperforms traditional methods, such as grid search and NSGA-II, as well as other mainstream RL baselines, in terms of new material discovery speed, design accuracy, and success rate, fully validating the feasibility and superiority of the proposed method.
This advantage is further strengthened by a closed-loop design-to-synthesis validation, demonstrating that \textsc{AIMatDesign} can reliably translate computational predictions into experimentally realizable materials.

\bmhead{Future Work} 
To focus on expanding to multi-objective and multi-scale design, incorporating additional domain constraints to enhance the algorithm’s reliability in structural stability and experimental feasibility.
Furthermore, integrating high-throughput experimental platforms and real-time feedback mechanisms will enable the development of an adaptive closed-loop design process, continuously refining model bias. 
Finally, expanding \textsc{AIMatDesign} to other advanced material domains, such as battery materials and high-entropy alloys, will further demonstrate its generality and scalability, laying a solid foundation for the next generation of intelligent materials design.

\subsection*{Acknowledgments}
This work was sponsored by the National Key Research and Development Program of China (No.2023YFB4606200), Key Program of Science and Technology of Yunnan Province (No.202302AB080020), Key Project of Shanghai Zhangjiang National Independent Innovation Demonstration Zone (No. ZJ2021-ZD-006). 

\subsection*{Author contributions}
\textbf{Yeyong Yu}: Writing - Original draft, Data curation, Software, Implementation, Methodology, Investigation, Formal analysis, Visualization. 
\textbf{Xilei Bian}: Data curation, BMGs Experimental validation and analysis, Writing - Review \& Editing. 
\textbf{Jie Xiong}: Data curation, Investigation, Writing - Review \& Editing. 
\textbf{Xing Wu}: Investigation, Writing - Review \& Editing. 
\textbf{Quan Qian}: Conceptualization, Methodology, Funding acquisition, Project administration, Supervision,
Writing - Review \& Editing.

\subsection*{Competing interests}
The authors declare that they have no conflicts of interest/competing interests.

\subsection*{Data and code availability} 
The data and source code that support the findings are available at \href{https://github.com/yuyouyu32/AIMatDesign}{https://github.com/yuyouyu32/AIMatDesign}.

\bibliographystyle{IEEEtran}
\bibliography{reference} 

\begin{thebibliography}{10}
\providecommand{\url}[1]{#1}
\csname url@samestyle\endcsname
\providecommand{\newblock}{\relax}
\providecommand{\bibinfo}[2]{#2}
\providecommand{\BIBentrySTDinterwordspacing}{\spaceskip=0pt\relax}
\providecommand{\BIBentryALTinterwordstretchfactor}{4}
\providecommand{\BIBentryALTinterwordspacing}{\spaceskip=\fontdimen2\font plus
\BIBentryALTinterwordstretchfactor\fontdimen3\font minus \fontdimen4\font\relax}
\providecommand{\BIBforeignlanguage}[2]{{%
\expandafter\ifx\csname l@#1\endcsname\relax
\typeout{** WARNING: IEEEtran.bst: No hyphenation pattern has been}%
\typeout{** loaded for the language `#1'. Using the pattern for}%
\typeout{** the default language instead.}%
\else
\language=\csname l@#1\endcsname
\fi
#2}}
\providecommand{\BIBdecl}{\relax}
\BIBdecl

\bibitem{field2001market}
F.~Field~III, J.~Clark, and M.~Ashby, ``Market drivers for materials and process development in the 21st century,'' \emph{MRS Bulletin}, vol.~26, no.~9, pp. 716--725, 2001.

\bibitem{capjon2004trial}
J.~Capjon, ``Trial-and-error-based innovation: Rapid materialisation as catalyser of perception and communication in design.'' 2004.

\bibitem{kang2022advances}
J.~Kang, X.~Zhang, and S.-H. Wei, ``Advances and challenges in dft-based energy materials design,'' \emph{Chinese Physics B}, vol.~31, no.~10, p. 107105, 2022.

\bibitem{saal2013materials}
J.~E. Saal, S.~Kirklin, M.~Aykol, B.~Meredig, and C.~Wolverton, ``Materials design and discovery with high-throughput density functional theory: the open quantum materials database (oqmd),'' \emph{Jom}, vol.~65, pp. 1501--1509, 2013.

\bibitem{maier2007combinatorial}
W.~F. Maier, K.~Stoewe, and S.~Sieg, ``Combinatorial and high-throughput materials science,'' \emph{Angewandte chemie international edition}, vol.~46, no.~32, pp. 6016--6067, 2007.

\bibitem{m2017mdts}
T.~M.~Dieb, S.~Ju, K.~Yoshizoe, Z.~Hou, J.~Shiomi, and K.~Tsuda, ``Mdts: automatic complex materials design using monte carlo tree search,'' \emph{Science and technology of advanced materials}, vol.~18, no.~1, pp. 498--503, 2017.

\bibitem{frazier2016bayesian}
P.~I. Frazier and J.~Wang, ``Bayesian optimization for materials design,'' \emph{Information science for materials discovery and design}, pp. 45--75, 2016.

\bibitem{chakraborti2004genetic}
N.~Chakraborti, ``Genetic algorithms in materials design and processing,'' \emph{International Materials Reviews}, vol.~49, no. 3-4, pp. 246--260, 2004.

\bibitem{liu2017materials}
Y.~Liu, T.~Zhao, W.~Ju, and S.~Shi, ``Materials discovery and design using machine learning,'' \emph{Journal of Materiomics}, vol.~3, no.~3, pp. 159--177, 2017.

\bibitem{ramprasad2017machine}
R.~Ramprasad, R.~Batra, G.~Pilania, A.~Mannodi-Kanakkithodi, and C.~Kim, ``Machine learning in materials informatics: recent applications and prospects,'' \emph{npj Computational Materials}, vol.~3, no.~1, p.~54, 2017.

\bibitem{menon2022generative}
D.~Menon and R.~Ranganathan, ``A generative approach to materials discovery, design, and optimization,'' \emph{ACS omega}, vol.~7, no.~30, pp. 25\,958--25\,973, 2022.

\bibitem{dan2020generative}
Y.~Dan, Y.~Zhao, X.~Li, S.~Li, M.~Hu, and J.~Hu, ``Generative adversarial networks (gan) based efficient sampling of chemical composition space for inverse design of inorganic materials,'' \emph{npj Computational Materials}, vol.~6, no.~1, p.~84, 2020.

\bibitem{lew2021encoding}
A.~J. Lew and M.~J. Buehler, ``Encoding and exploring latent design space of optimal material structures via a vae-lstm model,'' \emph{Forces in Mechanics}, vol.~5, p. 100054, 2021.

\bibitem{el2023vae}
K.~El-Awady, ``Vae for modified 1-hot generative materials modeling, a step towards inverse material design,'' \emph{arXiv preprint arXiv:2401.06779}, 2023.

\bibitem{guo2020semi}
K.~Guo and M.~J. Buehler, ``A semi-supervised approach to architected materials design using graph neural networks,'' \emph{Extreme Mechanics Letters}, vol.~41, p. 101029, 2020.

\bibitem{wang2021inverse}
Q.~Wang and L.~Zhang, ``Inverse design of glass structure with deep graph neural networks,'' \emph{Nature communications}, vol.~12, no.~1, p. 5359, 2021.

\bibitem{yu2024small}
Y.~Yu, J.~Xiong, X.~Wu, and Q.~Qian, ``From small data modeling to large language model screening: A dual-strategy framework for materials intelligent design,'' \emph{Advanced Science}, vol.~11, no.~45, p. 2403548, 2024.

\bibitem{zhang2021multi}
P.~Zhang, Y.~Qian, and Q.~Qian, ``Multi-objective optimization for materials design with improved nsga-ii,'' \emph{Materials today communications}, vol.~28, p. 102709, 2021.

\bibitem{ma2023comprehensive}
H.~Ma, Y.~Zhang, S.~Sun, T.~Liu, and Y.~Shan, ``A comprehensive survey on nsga-ii for multi-objective optimization and applications,'' \emph{Artificial Intelligence Review}, vol.~56, no.~12, pp. 15\,217--15\,270, 2023.

\bibitem{li2017deep}
Y.~Li, ``Deep reinforcement learning: An overview,'' \emph{arXiv preprint arXiv:1701.07274}, 2017.

\bibitem{sui2021deep}
F.~Sui, R.~Guo, Z.~Zhang, G.~X. Gu, and L.~Lin, ``Deep reinforcement learning for digital materials design,'' \emph{ACS Materials Letters}, vol.~3, no.~10, pp. 1433--1439, 2021.

\bibitem{brown2022deep}
N.~K. Brown, A.~P. Garland, G.~M. Fadel, and G.~Li, ``Deep reinforcement learning for engineering design through topology optimization of elementally discretized design domains,'' \emph{Materials \& Design}, vol. 218, p. 110672, 2022.

\bibitem{shah2021reinforcement}
T.~Shah, L.~Zhuo, P.~Lai, D.~La~Rosa-Moreno, F.~Amirkulova, P.~Gerstoft \emph{et~al.}, ``Reinforcement learning applied to metamaterial design,'' \emph{The Journal of the Acoustical Society of America}, vol. 150, no.~1, pp. 321--338, 2021.

\bibitem{karpovich2024deep}
C.~Karpovich, E.~Pan, and E.~A. Olivetti, ``Deep reinforcement learning for inverse inorganic materials design,'' \emph{npj Computational Materials}, vol.~10, no.~1, p. 287, 2024.

\bibitem{jia2024llmatdesign}
S.~Jia, C.~Zhang, and V.~Fung, ``Llmatdesign: Autonomous materials discovery with large language models,'' \emph{arXiv preprint arXiv:2406.13163}, 2024.

\bibitem{liu2024beyond}
Q.~Liu, M.~P. Polak, S.~Y. Kim, M.~Shuvo, H.~S. Deodhar, J.~Han, D.~Morgan, and H.~Oh, ``Beyond designer's knowledge: Generating materials design hypotheses via large language models,'' \emph{arXiv preprint arXiv:2409.06756}, 2024.

\bibitem{liashchynskyi2019grid}
P.~Liashchynskyi and P.~Liashchynskyi, ``Grid search, random search, genetic algorithm: a big comparison for nas,'' \emph{arXiv preprint arXiv:1912.06059}, 2019.

\bibitem{shahriari2015taking}
B.~Shahriari, K.~Swersky, Z.~Wang, R.~P. Adams, and N.~De~Freitas, ``Taking the human out of the loop: A review of bayesian optimization,'' \emph{Proceedings of the IEEE}, vol. 104, no.~1, pp. 148--175, 2015.

\bibitem{deb2002fast}
K.~Deb, A.~Pratap, S.~Agarwal, and T.~Meyarivan, ``A fast and elitist multiobjective genetic algorithm: Nsga-ii,'' \emph{IEEE transactions on evolutionary computation}, vol.~6, no.~2, pp. 182--197, 2002.

\bibitem{tian2024high}
Z.~Tian, Y.~Yang, S.~Zhou, T.~Zhou, K.~Deng, C.~Ji, Y.~He, and J.~S. Liu, ``High-dimensional bayesian optimization for metamaterial design,'' \emph{Materials Genome Engineering Advances}, vol.~2, no.~4, p. e79, 2024.

\bibitem{gao2019topology}
J.~Gao, H.~Xue, L.~Gao, and Z.~Luo, ``Topology optimization for auxetic metamaterials based on isogeometric analysis,'' \emph{Computer Methods in Applied Mechanics and Engineering}, vol. 352, pp. 211--236, 2019.

\bibitem{liao2024topological}
L.~Liao, S.~Yao, and Y.~Li, ``Topological optimization design of multi-material phononic crystals with floating projection constraints to achieve ultra-wide band gap,'' \emph{Composite Structures}, vol. 346, p. 118387, 2024.

\bibitem{muc2021introduction}
A.~Muc, ``Introduction to macroscopic optimal design in the mechanics of composite materials and structures,'' \emph{Journal of Composites Science}, vol.~5, no.~2, p.~36, 2021.

\bibitem{drugan2013designing}
M.~M. Drugan and A.~Nowe, ``Designing multi-objective multi-armed bandits algorithms: A study,'' in \emph{The 2013 international joint conference on neural networks (IJCNN)}.\hskip 1em plus 0.5em minus 0.4em\relax IEEE, 2013, pp. 1--8.

\bibitem{xiong2020machine2}
J.~Xiong, S.-Q. Shi, and T.-Y. Zhang, ``A machine-learning approach to predicting and understanding the properties of amorphous metallic alloys,'' \emph{Materials \& Design}, vol. 187, p. 108378, 2020.

\bibitem{lavalley2008logistic}
M.~P. LaValley, ``Logistic regression,'' \emph{Circulation}, vol. 117, no.~18, pp. 2395--2399, 2008.

\bibitem{xanthopoulos2013linear}
P.~Xanthopoulos, P.~M. Pardalos, T.~B. Trafalis, P.~Xanthopoulos, P.~M. Pardalos, and T.~B. Trafalis, ``Linear discriminant analysis,'' \emph{Robust data mining}, pp. 27--33, 2013.

\bibitem{lau2003online}
K.~Lau and Q.~Wu, ``Online training of support vector classifier,'' \emph{Pattern Recognition}, vol.~36, no.~8, pp. 1913--1920, 2003.

\bibitem{ying2015decision}
L.~Ying \emph{et~al.}, ``Decision tree methods: applications for classification and prediction,'' \emph{Shanghai archives of psychiatry}, vol.~27, no.~2, p. 130, 2015.

\bibitem{breiman2001random}
L.~Breiman, ``Random forests,'' \emph{Machine learning}, vol.~45, pp. 5--32, 2001.

\bibitem{natekin2013gradient}
A.~Natekin and A.~Knoll, ``Gradient boosting machines, a tutorial,'' \emph{Frontiers in neurorobotics}, vol.~7, p.~21, 2013.

\bibitem{chen2016xgboost}
T.~Chen and C.~Guestrin, ``Xgboost: A scalable tree boosting system,'' in \emph{Proceedings of the 22nd acm sigkdd international conference on knowledge discovery and data mining}, 2016, pp. 785--794.

\bibitem{prokhorenkova2018catboost}
L.~Prokhorenkova, G.~Gusev, A.~Vorobev, A.~V. Dorogush, and A.~Gulin, ``Catboost: unbiased boosting with categorical features,'' \emph{Advances in neural information processing systems}, vol.~31, 2018.

\bibitem{zhu2009multi}
J.~Zhu, H.~Zou, S.~Rosset, T.~Hastie \emph{et~al.}, ``Multi-class adaboost,'' \emph{Statistics and its Interface}, vol.~2, no.~3, pp. 349--360, 2009.

\bibitem{kramer2013k}
O.~Kramer and O.~Kramer, ``K-nearest neighbors,'' \emph{Dimensionality reduction with unsupervised nearest neighbors}, pp. 13--23, 2013.

\bibitem{ontivero2017fast}
M.~Ontivero-Ortega, A.~Lage-Castellanos, G.~Valente, R.~Goebel, and M.~Valdes-Sosa, ``Fast gaussian na{\"\i}ve bayes for searchlight classification analysis,'' \emph{Neuroimage}, vol. 163, pp. 471--479, 2017.

\bibitem{abbas2019multinomial}
M.~Abbas, K.~A. Memon, A.~A. Jamali, S.~Memon, and A.~Ahmed, ``Multinomial naive bayes classification model for sentiment analysis,'' \emph{IJCSNS Int. J. Comput. Sci. Netw. Secur}, vol.~19, no.~3, p.~62, 2019.

\bibitem{murphy2006naive}
K.~P. Murphy \emph{et~al.}, ``Naive bayes classifiers,'' \emph{University of British Columbia}, vol.~18, no.~60, pp. 1--8, 2006.

\bibitem{tharwat2016linear}
A.~Tharwat, ``Linear vs. quadratic discriminant analysis classifier: a tutorial,'' \emph{International Journal of Applied Pattern Recognition}, vol.~3, no.~2, pp. 145--180, 2016.

\bibitem{mcdonald2009ridge}
G.~C. McDonald, ``Ridge regression,'' \emph{Wiley Interdisciplinary Reviews: Computational Statistics}, vol.~1, no.~1, pp. 93--100, 2009.

\bibitem{ranstam2018lasso}
J.~Ranstam and J.~A. Cook, ``Lasso regression,'' \emph{Journal of British Surgery}, vol. 105, no.~10, pp. 1348--1348, 2018.

\bibitem{zou2005regularization}
H.~Zou and T.~Hastie, ``Regularization and variable selection via the elastic net,'' \emph{Journal of the Royal Statistical Society Series B: Statistical Methodology}, vol.~67, no.~2, pp. 301--320, 2005.

\bibitem{awad2015support}
M.~Awad, R.~Khanna, M.~Awad, and R.~Khanna, ``Support vector regression,'' \emph{Efficient learning machines: Theories, concepts, and applications for engineers and system designers}, pp. 67--80, 2015.

\bibitem{solomatine2004adaboost}
D.~P. Solomatine and D.~L. Shrestha, ``Adaboost. rt: a boosting algorithm for regression problems,'' in \emph{2004 IEEE international joint conference on neural networks (IEEE Cat. No. 04CH37541)}, vol.~2.\hskip 1em plus 0.5em minus 0.4em\relax IEEE, 2004, pp. 1163--1168.

\bibitem{hu2022ensemble}
M.~Hu, J.~H. Chion, P.~N. Suganthan, and R.~K. Katuwal, ``Ensemble deep random vector functional link neural network for regression,'' \emph{IEEE Transactions on Systems, Man, and Cybernetics: Systems}, vol.~53, no.~5, pp. 2604--2615, 2022.

\bibitem{hurst2024gpt}
A.~Hurst, A.~Lerer, A.~P. Goucher, A.~Perelman, A.~Ramesh, A.~Clark, A.~Ostrow, A.~Welihinda, A.~Hayes, A.~Radford \emph{et~al.}, ``Gpt-4o system card,'' \emph{arXiv preprint arXiv:2410.21276}, 2024.

\bibitem{mnih2015human}
V.~Mnih, K.~Kavukcuoglu, D.~Silver, A.~A. Rusu, J.~Veness, M.~G. Bellemare, A.~Graves, M.~Riedmiller, A.~K. Fidjeland, G.~Ostrovski \emph{et~al.}, ``Human-level control through deep reinforcement learning,'' \emph{nature}, vol. 518, no. 7540, pp. 529--533, 2015.

\bibitem{lillicrap2015continuous}
T.~P. Lillicrap, J.~J. Hunt, A.~Pritzel, N.~Heess, T.~Erez, Y.~Tassa, D.~Silver, and D.~Wierstra, ``Continuous control with deep reinforcement learning,'' \emph{arXiv preprint arXiv:1509.02971}, 2015.

\bibitem{fujimoto2018addressing}
S.~Fujimoto, H.~Hoof, and D.~Meger, ``Addressing function approximation error in actor-critic methods,'' in \emph{International conference on machine learning}.\hskip 1em plus 0.5em minus 0.4em\relax PMLR, 2018, pp. 1587--1596.

\bibitem{haarnoja2018soft}
T.~Haarnoja, A.~Zhou, K.~Hartikainen, G.~Tucker, S.~Ha, J.~Tan, V.~Kumar, H.~Zhu, A.~Gupta, P.~Abbeel \emph{et~al.}, ``Soft actor-critic algorithms and applications,'' \emph{arXiv preprint arXiv:1812.05905}, 2018.

\bibitem{schulman2017proximal}
J.~Schulman, F.~Wolski, P.~Dhariwal, A.~Radford, and O.~Klimov, ``Proximal policy optimization algorithms,'' \emph{arXiv preprint arXiv:1707.06347}, 2017.

\bibitem{hartigan1979algorithm}
J.~A. Hartigan and M.~A. Wong, ``Algorithm as 136: A k-means clustering algorithm,'' \emph{Journal of the royal statistical society. series c (applied statistics)}, vol.~28, no.~1, pp. 100--108, 1979.

\bibitem{lewis2020retrieval}
P.~Lewis, E.~Perez, A.~Piktus, F.~Petroni, V.~Karpukhin, N.~Goyal, H.~K{\"u}ttler, M.~Lewis, W.-t. Yih, T.~Rockt{\"a}schel \emph{et~al.}, ``Retrieval-augmented generation for knowledge-intensive nlp tasks,'' \emph{Advances in neural information processing systems}, vol.~33, pp. 9459--9474, 2020.

\end{thebibliography}




\clearpage
\appendix 

\section{Supplementary Experimental Results}
\label{A:supplementary_results}

\subsection{Classification Results}
\label{A:cls_results}

we present the performance comparison of classification models based on 5-fold cross-validation. Detailed metrics are provided in Table~\ref{tab:cls_results}. The table compares the performance of multiple classification models across AUC, Precision, Recall, and F1 score. Overall, both CatBoost and RF (Random Forest) outperformed the other models. While models such as SVC, XGBoost, and AdaBoost also demonstrated strong performance in certain metrics, their overall performance was slightly lower than that of CatBoost and RF.

In this study, we use four main performance evaluation metrics: Precision, Recall, and F1 score, AUC. These are defined as follows:
\begin{itemize}
	\item \textbf{Precision}: Precision is the ratio of correctly predicted positives to the total predicted positives. It indicates how accurate the positive predictions are.
	\begin{equation}
	\text{Precision} = \frac{TP}{TP + FP}
	\end{equation}
	\item \textbf{Recall}: Recall is the ratio of correctly predicted positives to all actual positives. It reflects the model's ability to detect all relevant cases.
	\begin{equation}
	\text{Recall} = \frac{TP}{TP + FN}
	\end{equation}
	\item \textbf{F1 Score}: The F1 score is the harmonic mean of Precision and Recall, balancing both metrics, and is useful for imbalanced class distributions.
	\begin{equation}
		F1 = 2 \times \frac{\text{Precision} \times \text{Recall}}{\text{Precision} + \text{Recall}}
	\end{equation}
	\item \textbf{AUC (Area Under the ROC Curve)}: AUC measures the separability of the model. It ranges from 0 to 1, with 1 indicating perfect classification and 0.5 indicating no discriminative power.
	\begin{equation}
	\text{AUC} = \int_{0}^{1} \text{TPR}(x) \, dx
	\end{equation}
\end{itemize}

\begin{table}[ht]
	\centering
	\caption{The 5-fold cross-validation performance comparison of various classification models based on AUC, Precision, Recall, and F1 score metrics.}
	\label{tab:cls_results}
	\setlength{\tabcolsep}{4pt} 
	\resizebox{\textwidth}{!}{%
	\renewcommand{\arraystretch}{1.25} 
	\begin{tabular}{lcccccccccccccc}
		\toprule
		 &
		  \textbf{LR} &
		  \textbf{SVC} &
		  \textbf{RF} &
		  \textbf{GBM} &
		  \textbf{AdaBoost} &
		  \textbf{KNN} &
		  \textbf{XGBoost} &
		  \textbf{DT} &
		  \textbf{GNB} &
		  \textbf{MNB} &
		  \textbf{BNB} &
		  \textbf{LDA} &
		  \textbf{QDA} &
		  \textbf{CatBoost} \\ \midrule
		  \textbf{AUC} & 0.79 & \textbf{0.95} & \textbf{0.95} & \textbf{0.95} & 0.92 & \textbf{0.95} & \textbf{0.95} & 0.91 & 0.59 & 0.7 & 0.79 & 0.78 & 0.58 & \textbf{0.95} \\
		  \textbf{Precision} & 0.5 & 0.88 & \textbf{0.96} & 0.94 & 0.84 & 0.83 & 0.94 & 0.86 & 0.68 & 0.4 & 0.54 & 0.47 & 0.83 & 0.95 \\
		  \textbf{Recall} & 0.79 & \textbf{0.93} & 0.91 & 0.91 & 0.88 & 0.94 & 0.92 & 0.86 & 0.19 & 0.67 & 0.75 & 0.79 & 0.17 & 0.92 \\
		  \textbf{F1 score} & 0.61 & 0.91 & \textbf{0.93} & 0.93 & 0.86 & 0.89 & 0.93 & 0.86 & 0.3 & 0.5 & 0.63 & 0.59 & 0.28 & \textbf{0.94} \\ \bottomrule
		\end{tabular}%
	}
\end{table}

\subsection{Regression Results}
\label{A:regression_results}
We present the performance comparison of regression models across several evaluation metrics. Detailed metrics are provided in Table~\ref{tab:regression_results}. 
The table compares the performance of multiple regression models across RMSE, $R^2$, and MAPE. 
Overall, the edRVFL model outperforms the other models. While models such as Ridge, Lasso, and XGBoost showed strong performance in specific metrics, edRVFL consistently performed better across multiple key metrics.

In this study, we use three main performance evaluation metrics: RMSE, $R^2$, and MAPE. These are defined as follows:
\begin{itemize}
	\item \textbf{RMSE}: RMSE measures the average magnitude of the error, with a lower RMSE indicating better model performance. It is defined as:
	\begin{equation}
	\text{RMSE} = \sqrt{\frac{1}{n} \sum_{i=1}^{n} (y_i - \hat{y}_i)^2}
	\end{equation}
	where \(y_i\) is the actual value, \(\hat{y}_i\) is the predicted value, and \(n\) is the number of data points.

	\item $\bm{R^2$}: $R^2$ indicates how well the model explains the variance of the data. A value closer to 1 indicates a better fit:
	\begin{equation}
	R^2 = 1 - \frac{\sum_{i=1}^{n} (y_i - \hat{y}_i)^2}{\sum_{i=1}^{n} (y_i - \bar{y})^2}
	\end{equation}
	where \(y_i\) is the actual value, \(\hat{y}_i\) is the predicted value, and \(\bar{y}\) is the mean of the actual values.

	\item \textbf{MAPE}: MAPE measures the average percentage difference between predicted and actual values, providing an indication of the relative prediction error:
	\begin{equation}
		\text{MAPE} = \frac{100}{n} \times \sum_{i=1}^{n} \frac{y_i - \hat{y}_i}{y_i}
		\end{equation}
	where \(y_i\) is the actual value, \(\hat{y}_i\) is the predicted value, and \(n\) is the number of data points.
\end{itemize}

\begin{table}[ht]
	\centering
	\caption{The 5-fold cross-validation performance comparison of various regression models based on RMSE, $R^2$, and MAPE metrics.}
	\label{tab:regression_results}
	\resizebox{\textwidth}{!}{%
	\renewcommand{\arraystretch}{1.25} 
	\begin{tabular}{p{0.1\textwidth}llcccccccccc} \toprule
		\multicolumn{2}{l}{\textbf{}}                           & \textbf{Ridge} & \textbf{Lasso} & \textbf{ElasticNet} & \textbf{SVR} & \textbf{RF} & \textbf{GBM} & \textbf{AdaBoost} & \textbf{KNN} & \textbf{XGBoost} & \textbf{edRVFL} \\ \midrule
		\multirow{3}{*}{\textit{\textbf{$\bm{D_{\text{max}}}$(mm)}}}       & \textbf{RMSE} & 0.14 & 0.14 & 0.14 & 0.11 & 0.1 & 0.1 & 0.12 & 0.11 & 0.1 & \textbf{0.07} \\
												& \textbf{$\bm{R^2}$}   & 0.19 & 0.16 & 0.17 & 0.44 & 0.58 & 0.55 & 0.4 & 0.46 & 0.55 & \textbf{0.8} \\
												& \textbf{MAPE} & 124.95 & 136.07 & 127.6 & 132.66 & 75.68 & 81.4 & 137.89 & 98.69 & 75.09 & \textbf{49.27} \\ \hline
		\multirow{3}{*}{\textit{\textbf{Tg(K)}}}        & \textbf{RMSE} & 0.06 & 0.09 & 0.07 & 0.07 & 0.04 & 0.04 & 0.09 & 0.05 & 0.04 & \textbf{0.03}   \\
												& \textbf{$\bm{R^2}$}   & 0.91 & 0.81 & 0.87 & 0.88 & 0.96 & 0.97 & 0.79 & 0.94 & 0.96 & \textbf{0.98} \\ 
												& \textbf{MAPE} & 15.97 & 29.75 & 22.28 & 24.72 & 8.64 & 7.22 & 35 & 7.94 & 7.88 & \textbf{5.26}   \\ \hline
		\multirow{3}{*}{\textit{\textbf{Tl(K)}}}        & \textbf{RMSE}  & 0.05 & 0.08 & 0.07 & 0.06 & 0.05 & 0.05 & 0.09 & 0.05 & 0.05 & \textbf{0.02} \\
												& \textbf{$\bm{R^2}$}   & 0.95 & 0.87 & 0.92 & 0.93 & 0.96 & 0.96 & 0.85 & 0.95 & 0.95 & \textbf{0.99} \\
												& \textbf{MAPE} & 17.64 & 42.37 & 34.42 & 34.58 & 12.82 & 9.68 & 61.54 & 10.08 & 12.22 & \textbf{5.26} \\ \hline
		\multirow{3}{*}{\textit{\textbf{Tx(K)}}}         & \textbf{RMSE} & 0.07 & 0.1 & 0.08 & 0.08 & 0.06 & 0.05 & 0.1 & 0.05 & 0.05 & \textbf{0.03} \\
												& \textbf{$\bm{R^2}$}   & 0.91 & 0.84 & 0.88 & 0.9 & 0.94 & 0.95 & 0.83 & 0.96 & 0.95 & \textbf{0.98} \\
												& \textbf{MAPE} & 13.27 & 18.03 & 15.36 & 15.81 & 7.04 & 6.27 & 21.71 & 6.55 & 7.55 & \textbf{3.83} \\ \hline
		\multirow{3}{*}{\textit{\textbf{$\bm{\sigma_Y}$(MPa)}}}    & \textbf{RMSE} & 0.11 & 0.12 & 0.12 & 0.09 & 0.08 & 0.08 & 0.1 & 0.09 & 0.09 & \textbf{0.03} \\
												& \textbf{$\bm{R^2}$}   & 0.21 & 0.11 & 0.12 & 0.49 & 0.57 & 0.61 & 0.39 & 0.42 & 0.53 & \textbf{0.91}   \\
												& \textbf{MAPE} & 50.02 & 64.44 & 63.19 & 38.47 & 31.79 & 27.68 & 42.98 & 30.8 & 33.95 & \textbf{10.3} \\ \hline
		\multirow{3}{*}{\textit{\textbf{$\bm{E}$(GPa)}}} & \textbf{RMSE} & 0.09 & 0.12 & 0.1 & 0.1 & 0.09 & 0.08 & 0.11 & 0.08 & 0.08 & \textbf{0.05}   \\
												& \textbf{$\bm{R^2}$}   & 0.74 & 0.57 & 0.66 & 0.69 & 0.74 & 0.8 & 0.6 & 0.78 & 0.8 & \textbf{0.89} \\
												& \textbf{MAPE} & 23.76 & 52.97 & 34.16 & 41.64 & 19.2 & 22.64 & 82.36 & 19.18 & 22.46 & \textbf{15.62} \\ \hline
		\multirow{3}{*}{\textit{\textbf{$\bm{\varepsilon(\%)}$}}}          & \textbf{RMSE} & 0.12 & 0.13 & 0.13 & 0.1 & 0.1 & 0.1 & 0.13 & 0.12 & 0.1 & \textbf{0.04} \\
												& \textbf{$\bm{R^2}$}   & 0.44 & 0.34 & 0.4 & 0.6 & 0.62 & 0.64 & 0.41 & 0.52 & 0.65 & \textbf{0.92} \\
												& \textbf{MAPE} & 233.3 & 240.54 & 226.74 & 268.18 & 172.04 & 181.92 & 235.6 & 192.8 & 166.99 & \textbf{84.04} \\ \bottomrule
		\end{tabular}%
}
\end{table}

\section{Implementation Details}
\label{A:implementation_details}

\bmhead{Training} 
ML models employed stratified 5-fold cross-validation (StratifiedKFold) for parameter optimization, with grid search (GridSearchCV) evaluating model performance across predefined hyperparameter spaces. For classification models, the area under the ROC curve (AUC) was used as the evaluation metric, while for regression models, the $R^2$ score was utilized. The cross-validation process maintained class distribution consistency and was accelerated using 12-thread parallel computing. RL models were configured with a batch size of 512 and a total of 100,000 training steps, incorporating Prioritized Experience Replay (PER) to optimize experience sampling efficiency.

\bmhead{Inference} 
LLM inference utilized standard API parameters: temperature coefficient (0.7) controlled generation diversity, nucleus sampling threshold ($top_p=0.95$) ensured 95\% probability mass coverage, and maximum generation length was constrained to 4096 tokens. API calls implemented exponential backoff retry mechanisms (maximum 3 attempts).

\bmhead{Evaluation} 
\textsection~\ref{rl_results} employed a phased evaluation protocol, where each epoch began with the random selection of one of the 35 exploration bases, followed by the selection of a component as $s_0$. Each epoch consisted of up to 128 iterative steps, with early termination if stopping criteria were met, and the entire trial spanned 1,000 training epochs. \textsection~\ref{design_results} involved 100 evaluation epochs for each exploration base, where each base commenced with a randomly selected component as $s_0$.

\bmhead{Hardware and System Configuration} 
We use 2 NVIDIA RTX V100 GPUs with 128GB of memory for training and a single V100 GPU for inference. The system operates on Linux version 4.14.105-1-tlinux3-0013. Software stack included: Python 3.8, PyTorch 2.0.1 with CUDA 11.8 and cuDNN 8.6.0 acceleration.

\section{Prompt Templates}
\label{A:prompt_templates}

The prompt templates in Table~\ref{prompt:knowledge_reward} are used to evaluate the Knowledge-Based Reward (KBR). 
In these templates, the \textbf{\{rule\}} section contains the criteria derived by LLMs based on relevant materials science knowledge obtained from both provided papers and web searches. These rules provide clear evaluation standards, and LLMs assess data points according to them, ensuring that the evaluation process is scientifically grounded and consistent. 
This approach allows the model not only to rely on existing experimental data and literature but also to automatically incorporate multiple knowledge sources, leading to more accurate and practical reward evaluations.

\begin{table}[ht]
\centering
\resizebox{.98\textwidth}{!}{
\begin{tabular}{l}
\toprule
\textbf{Prompt for Knowledge-Based Reward} \\
\hline 
	You are an expert in materials science with extensive experience in Bulk Metallic Glass (BMG) composition, \\ performance, and experimental validation. \\
	You can objectively assess the potential of BMG compositions using scientific principles and experimental data.\\
	\\
	Given the following selection criteria (RULE) and performance data of similar BMGs (Similar Real BMGs), \\
	evaluate the provided data point (DATA) to determine its suitability for experimental validation. \\
	Assign a reward value between -1 and 1 to guide the reinforcement learning (RL) model's Knowledge-Base Reward.\\ 
	\\
Provide a detailed reasoning process to ensure scientific accuracy:\\
1. Review and understand the selection criteria (RULE), identifying key indicators and requirements.\\
2. Compare with similar BMGs, analyzing performance characteristics and experimental outcomes as benchmarks.\\
3. Evaluate the provided data point (DATA) against the selection criteria and reference data, assigning a reward value\\ and justifying your reasoning.\\
\\
\textbf{RULE:} \\
\{rule\}\\
\\
\textbf{Similar Real BMGs:}\\
\{similar\_real\_bmg\}\\
\\
\textbf{DATA:}\\
\{data\}\\
\\
The reward value should range from -1 to 1, where 1 indicates high experimental value and alignment with BMG knowledge,\\ and -1 indicates significant deviation from the criteria. \\
Output the evaluation result in the following format: \\
\{\\
\qquad	``reward": Data point's reward value, \lbrack -1, 1\rbrack, rounded to two decimal places,\\
\qquad	``reason": Brief explanation of the assigned value\\
\}\\
Now please start evaluating the data points (DATA) and give the award value and reason for the evaluation. \\
Please note that the final evaluation results need to be output in JSON format to ensure that the format is correct.\\
\bottomrule
\end{tabular}
}
\caption{Prompt Template for Knowledge-Based Reward (KBR)}
\label{prompt:knowledge_reward}
\end{table}

The prompt templates in Table~\ref{prompt:variance_refinement} and Table~\ref{prompt:correlation_refinement} are used to perform Automatic Model Refinement (AMR), focusing on feature engineering to optimize the ML model. 
In these templates, the \textbf{\{knowledge\}} section utilizes retrieval-augmented generation (RAG)~\citep{lewis2020retrieval} techniques to extract relevant domain knowledge from a knowledge base, providing a scientific basis for the feature selection process. Meanwhile, the \textbf{\{Candidate Features\}} list, sourced from~\citep{xiong2020machine2}, includes various atomic-level computed features. The feature selection process follows a hierarchical approach, starting with broad feature categories and progressively narrowing down to specific features, ensuring that the final selected features effectively improve the model's predictive power and consistency.

In the \textbf{Variance-Based Refinement}, the template focuses on feature selection to reduce the fluctuation caused by high variance predictions, thereby enhancing the model's stability. In contrast, \textbf{Correlation-Based Refinement} aims to reduce the prediction discrepancies between the reinforcement learning model and the machine learning model, enhancing consistency between the two. By combining RAG and hierarchical filtering, the model is able to more accurately select the most valuable features for performance optimization from a large pool of candidate features, thereby improving both prediction accuracy and consistency.
\begin{table}[ht]
	\centering
	\resizebox{.98\textwidth}{!}{
	\begin{tabular}{l}
	\toprule
	\textbf{Prompt for Variance-Based Refinement} \\
	\hline 
	You are an expert in machine learning modeling for Bulk Metallic Glass (BMG), with in-depth knowledge of material \\composition, performance, and machine learning applications in materials science. \\
	You are able to accurately \\analyze the current model's issues and optimize model performance through feature engineering. \\
	\\
	Currently, the Guiding Model (regression model) exhibits high prediction variance (\{pred\_var\}) when predicting the
	\\
	\{performance\} of similar BMG \{composition\}, resulting in unstable predictions. To improve model \\
	performance, you need to select 1-3 new features from the provided candidate features and retrain the Guiding 
	Model \\(regression model) to help reduce prediction variance when predicting the \{performance\} of similar BMG compositions.\\
	\\
	Please follow these steps for feature selection:\\
	1. Analyze the current state of the Guiding Model, including the features used and the potential reasons for high \\ prediction variance, to identify areas for improvement.\\
	2. Evaluate each candidate feature, considering its correlation with the current high-variance BMG compositions and \\performance, data quality, and its potential impact on model prediction ability.\\
	3. Based on the evaluation of the model's improvement direction and candidate features, select the 1-3 most \\promising features and provide a brief explanation of why these features were chosen.\\
	\\
	\textbf{Reference Knowledge:}\\
	\{knowledge\}\\
	\\

	\textbf{Guiding Model Status:}\\
	\{model\_status\}\\
	\\
	\textbf{Candidate Features:}\\
	\{candidate\_features\}\\
	\\
	When selecting features, focus on identifying those that can effectively reduce instability in high-variance \\predictions or provide additional explanatory power, as well as those that correlate with the target performance,\\ {performance}. Finally, output the selected features and reasons in the following format:\\
	\{\\
	\qquad	``selected\_features": \lbrack ``feature1", ... \rbrack,\\
	\qquad	``reason": ``reason for selecting these features"\\
	\}\\
	Please start evaluating the candidate features and provide a detailed explanation of the selected features.\\ Ensure the final output meets the requirements and is returned in JSON format.\\
	\bottomrule
	\end{tabular}
	}
	\caption{Prompt Template for Variance-Based Refinement}
\label{prompt:variance_refinement}
\end{table}

\begin{table}[ht]
	\centering
	\resizebox{.98\textwidth}{!}{
	\begin{tabular}{l}
	\toprule
	\textbf{Prompt for Correlation-Based Refinement} \\
	\hline 
	You are an expert in machine learning modeling for Bulk Metallic Glass (BMG), with in-depth knowledge of material \\composition, performance, and machine learning applications in materials science. \\
	You are able to accurately \\analyze the current model's issues and optimize model performance through feature engineering. \\
	\\
	Currently, there is a significant divergence between the reward curve ($R_f$) provided by the Guiding Model (ML model) \\
	and the state value curve ($V_f$) provided by the Explore Model (RL model) when predicting the performance related \\to the \{composition\} composition, with a Pearson correlation coefficient of \{person\_cor\}. \\This indicates that the two models predict the same composition differently, leading to inconsistencies in their judgments.\\
	\\
	To improve the prediction consistency and performance of the models, you need to select 1-3 new features from the\\ provided candidate features to retrain the Guiding Model (regression model) to enhance the alignment between the \\machine learning model and the reinforcement learning model.\\
	\\
	Please follow these steps for feature selection:\\
	1. Analyze the current state of the Guiding Model, including the features used and the potential reasons for the low\\ 
	Pearson correlation between the reward curve ($R_f$) and the state value curve ($V_f$), and identify areas for improvement.\\
	2. Evaluate each candidate feature, considering its potential relationship with the current inconsistency in predictions, \\and assess whether adding the feature will improve the machine learning model's performance, \\helping it align with the reinforcement learning model.\\
	3. Select the most optimal features and justify your choice by considering the direction of model improvement and\\ the evaluation of candidate features. Select the 1-3 most promising features and briefly explain the rationale \\behind these selections.\\
	\\
	\textbf{Reference Knowledge:}\\
	\{knowledge\}\\
	\\
	\textbf{Guiding Model Status:}\\
	\{model\_status\}\\
	\\
	\textbf{Candidate Features:}\\
	\{candidate\_features\}\\
	\\
	When selecting features, focus on identifying those that can effectively reduce the inconsistency between the ML and\\ 
	RL models, provide additional explanatory power, and show significant correlation with the \{composition\}. \\Finally, output the selected features and their reasoning in the following format:\\
	\{\\
	\qquad	``selected\_features": \lbrack ``feature1", ... \rbrack,\\
	\qquad	``reason": ``reason for selecting these features"\\
	\}\\
	Please begin evaluating the candidate features and provide detailed explanations of the selected features.\\ Ensure the final output meets the requirements and is returned in JSON format.\\

	\bottomrule
	\end{tabular}
	}
	\caption{Prompt Template for Correlation-Based Refinement}
	\label{prompt:correlation_refinement}
\end{table}

\section{Open Access and Licensing}
\label{A:open_access}

The code used in this study is released under the Apache 2.0 License. The associated code repository is publicly available for use, modification, and distribution in compliance with the terms of the Apache 2.0 License.

The dataset used in this research is shared under the Creative Commons Attribution-NonCommercial 4.0 International (CC BY-NC 4.0) license. This dataset is available for non-commercial use and can be redistributed and modified under the terms specified by the license.

The code and dataset are provided in the supplementary files and will be made publicly available via open-source links upon acceptance of the paper. Detailed access instructions and relevant links will be included in the final version of the paper.

\end{document}